\documentclass[10pt,twocolumn,letterpaper]{article}

\usepackage[pagenumbers]{cvpr}      
\usepackage{adjustbox}
\usepackage{multirow}
\usepackage{makecell}
\usepackage{caption}

\usepackage[most]{tcolorbox}
\usepackage{dblfloatfix}
\usepackage{enumitem}
\usepackage{microtype}
\usepackage{lipsum} 
\usepackage{xspace}
\usepackage{upgreek}
\usepackage[utf8]{inputenc}

\definecolor{PromptBlue}{HTML}{1F4E79}
\definecolor{PromptBack}{HTML}{F2F6FB}

\newcommand{\LMInBox}{\fontfamily{lmr}\selectfont}
\newcommand{\SelfReflectionTitle}{Self-Reflection Prompt Template} 
\newtcolorbox{SelfReflectionBox*}[1][]{%
  selfreflectionstyle,
  float*,
  floatplacement=tbp,
  width=\textwidth,
  before=\vspace{2mm},
  after=\vspace{2mm},
  #1
}

\tcbset{
  selfreflectionstyle/.style={
    enhanced, breakable,
    colback=PromptBack, colframe=PromptBlue, boxrule=0.9pt, arc=2mm,
    left=3mm, right=3mm, top=2mm, bottom=2mm,
    fonttitle=\bfseries\large\LMInBox, 
    fontupper=\normalsize\LMInBox, 
    fontlower=\normalsize\LMInBox,
    coltitle=white,
    title=\SelfReflectionTitle,
    attach boxed title to top left={xshift=0mm,yshift*=-\tcboxedtitleheight/2},
    boxed title style={
      enhanced, colback=PromptBlue, colframe=PromptBlue,
      boxrule=0pt, arc=2mm, top=2pt, bottom=2pt,
      left=2.5mm, right=2.5mm
    }
  }
}

\newtcolorbox{SelfReflectionBox}{ selfreflectionstyle }

\setlist[itemize]{topsep=4pt,itemsep=2pt,left=6mm}
\setlist[enumerate]{topsep=4pt,itemsep=2pt,left=6mm}

\newcommand{\state}{s_t}
\newcommand{\obs}{o_t}

\newcommand{\loc}{l_t}
\newcommand{\mem}{\mathcal{M}_t}
\newcommand{\fronts}{\mathcal{F}_t}
\newcommand{\Ka}{K_\mathrm{a}}

\newcommand{\newpolicy}{\pi^{*}_{\theta}}
\newcommand{\retr}{\mathcal{R}_t}
\newcommand{\expr}{\mathcal{T}_{seen}}
\newcommand{\selfron}{f_t^{(a_t)}}
\newcommand{\bvf}{\mathcal{F}_t^{\text{BVF}}}
\newcommand{\cvf}{\mathcal{F}_t^{\text{CVF}}}

\definecolor{cvprblue}{rgb}{0.21,0.49,0.74}
\usepackage[pagebackref,breaklinks,colorlinks,allcolors=cvprblue]{hyperref}

\def\confName{CVPR}
\def\confYear{2026}

\newcommand{\explorer}{\texttt{\textbf{ReEXplore}}\xspace}

\newcommand{\llm}{LLM-Match\xspace}
\newcommand{\llmspl}{LLM-Match$\times$SPL\xspace}
\newcommand{\qwen}{Qwen2.5-VL-7B-Instruct\xspace}
\newcommand{\gpt}{GPT-4o\xspace}

\title{\explorer: Improving MLLMs for Embodied Exploration with Contextualized Retrospective Experience Replay}

\author{
\textbf{Gengyuan Zhang}$^{1,2}$\thanks{Equal contribution},
\textbf{Mingcong Ding}$^{3}$\footnotemark[1],
\textbf{Jingpei Wu}$^{1,2}$,
\textbf{Routong Liao}$^{1,2}$,
\textbf{Volker Tresp}$^{1,2}$\\[6pt]
$^{1}$LMU Munich,
$^{2}$Munich Center for Machine Learning,
$^{3}$TU Munich\\[6pt]
{\tt gengyuan.zhang@lmu.de \qquad\ mingcong.ding@tum.de}
}

\begin{document}

\twocolumn[{
\renewcommand\twocolumn[1][]{#1}
\maketitle
\begin{center}

\includegraphics[width=0.96\textwidth]{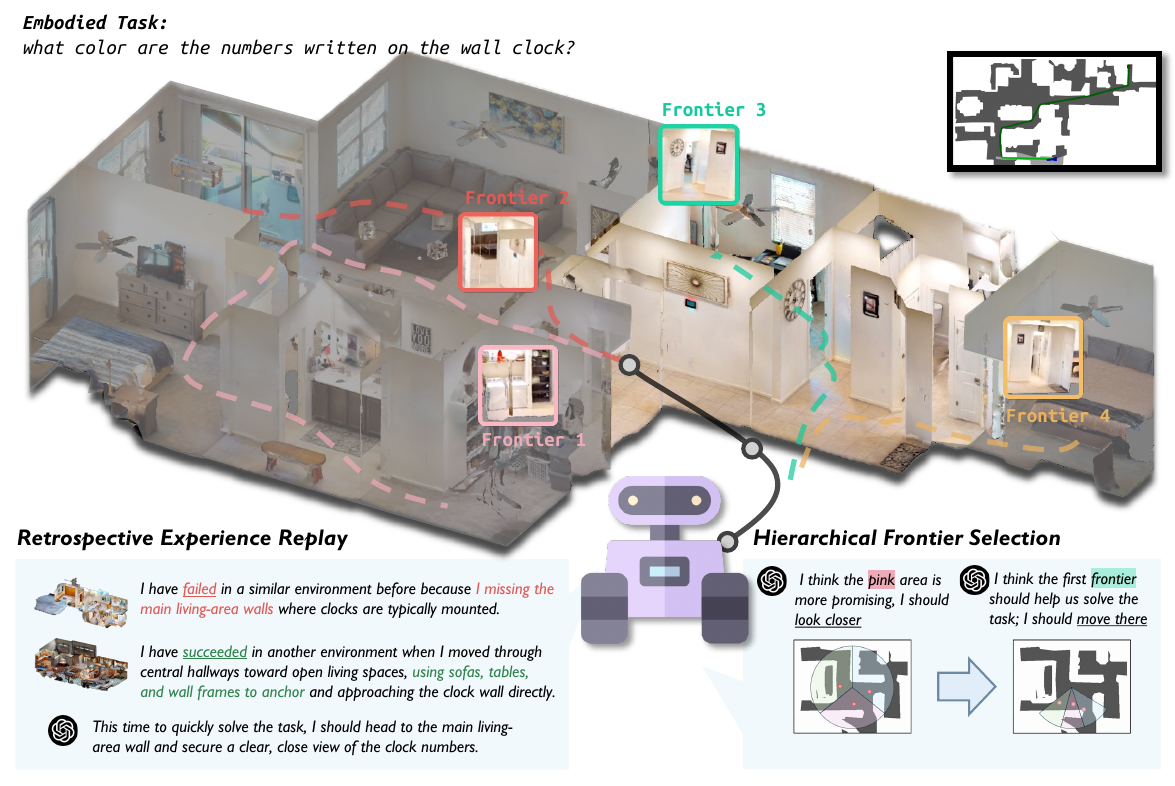}
\captionof{figure}{We introduce \explorer, a training-free framework that strengthens MLLM-based embodied agents for frontier-based exploration. Prior MLLM agents suffer from two core limitations: (1) dependence on stale pre-trained knowledge, and (2) difficulty distinguishing and ranking visually similar frontier candidates in a large action space. \explorer overcomes these challenges through \emph{Retrospective Experience Replay}, which injects distilled experiences from previous trials directly at inference time, and \emph{Hierarchical Frontier Selection}, which decomposes frontier space into coarse-to-fine decisions for more reliable and efficient exploration.}
\label{fig:teaser}
\end{center}
}]

\maketitle

\renewcommand{\thefootnote}{*}
\footnotetext{Equal Contribution}

\renewcommand{\thefootnote}{\arabic{footnote}}

\begin{abstract}
Embodied exploration is a target-driven process that requires embodied agents to possess fine-grained perception and knowledge-enhanced decision making.
While recent attempts leverage MLLMs for exploration due to their strong perceptual and reasoning abilities, we find that MLLM-based embodied agents remain suboptimal in exploring new environments: (i) they rely on profound but \emph{stale} pre-trained knowledge, (ii) training-based approaches such as imitation learning or reinforcement learning are expensive for long-horizon tasks with sparse outcome rewards, and (iii) frontier-based exploration yields a large, visually nuanced action space that is difficult for MLLMs to make reliable decisions.
We address these challenges with \explorer, a training-free framework that performs \emph{retrospective experience replay} to inject distilled, abstract experience at inference time, and \emph{hierarchical frontier selection} to decompose frontier ranking into coarse-to-fine decisions.
Our approach enables robust, traceable, and efficient exploration. Across multiple embodied exploration benchmarks, \explorer yields great improvements over strong MLLM baselines, up to $3\times$ higher performance in both success rate and in navigation efficiency under open-source backbones.

\end{abstract}

\section{Introduction}
\label{sec:intro}

Embodied agents operate in the physical world and must actively explore their surroundings to accomplish a series of tasks such as navigation, object localization, and embodied question answering~\cite{das2018embodied,fung2025embodied}.  For example, a household robot can search an entire flat to retrieve a remote control misplaced under a sofa; an aerial drone can navigate a collapsed building to locate survivors, or an inspector robot autonomously scan industrial facilities~\cite{li2025regnav}.  In such scenarios, effective exploration requires close coordination between perception, semantic reasoning, and decision making. This means that an agent must understand its observations, relate them to the task, and choose where to explore next.

Multimodal large language models (MLLMs) have recently emerged as promising ``eyes and brains'' for embodied agents~\cite{lin2023advances}.  Their joint visual–linguistic representations allow agents to reason over both the semantics of the environment and the task objective~\cite{wang2025multimodal}.  Compared with traditional frontier-based exploration~\cite{chen2023not, majumdar2024openeqa}, which greedily maximizes short-term information gain, MLLM-based agents can incorporate world knowledge and semantic priors to guide frontier selection\cite{wald2020learning}.  This enables a richer, more goal-directed exploration strategy in open environments. 

However, early attempts to deploy MLLMs~\cite{ren2024explore,yang20253dmem3dscenememory,majumdar2024openeqa,shridhar_alfworld_2021,zhao2025cityeqa,zhu2023excalibur} for embodied exploration expose several fundamental limitations.  
First, MLLMs are pre-trained on web-scale data and are not tailored to a specific environment; without real-world grounding, they may propose actions that appear plausible linguistically but are practically infeasible~\cite{ahn2022can}.  
Second, most embodied agents~\cite{majumdar2024openeqa,yang20253d} explore the environments in stateless manner and do not leverage accumulated observations or long-term cross-episode experience, leading to inefficient and repetitive exploration~\cite{lin2023advances}.  
Third, frontier-based exploration presents a large and visually confounding action space~\cite{kolve2017ai2}, where frontiers often number in the tens or hundreds and have merely subtle visual differences~\cite{gordon2018iqa} that make direct MLLM ranking unreliable and computationally heavy~\cite{meng2025ssr}.  

Together, these issues highlight the need for \emph{experience-driven} exploration that goes beyond stale pre-training knowledge and supports adaptive, context-aware decision making.
Previously, Reinforcement learning (RL) offers a parametric way to learn from experience~\cite{zhao2025embodied,zhu2023excalibur}, but is \emph{expensive} for embodied exploration tasks.  
Exploration tasks are long-horizon with hard-to-define intermediate rewards; its success or failure could be attributed to different steps; physical or simulated rollouts with MLLMs are resource-intensive~\cite{zhou2025reinforced}; and reward signals remain sparse~\cite{stranghoner2025share}.  
Even in simulation, RL exhibits severe sample inefficiency~\cite{wang2022learning}, making it impractical for training MLLMs.  

These limitations motivate a research question:  
\emph{Can we equip an MLLM-based agent with reusable exploration experience in a training-free, non-parametric manner?}  
As humans explores new environments and learn from their success or failure with a hindsight and reuse them in future attempts,
similarly an embodied agent should benefit from abstracted, retrospective experience that can be plugged into its policy on the fly.

To this end, we propose \explorer (\textbf{Re}trospective \textbf{EX}perience-based \textbf{EXplor}ation), a training-free framework that enables MLLMs to perform dynamic, adaptive exploration in unseen environments.  
Our approach offers multiple benefits: it avoids data-heavy training, eliminates reward design, and supports rapid adaptation to new scenarios.  
Specifically, \explorer (i) retrospectively abstracts completed exploration trajectories into concise, transferable experiences and replays them during future exploration, and (ii) hierarchically selects frontiers by decomposing the action space and using MLLM reasoning in a coarse-to-fine manner.  
As shown in ~\cref{fig:teaser}, this two-fold strategy allows the agent to reuse prior successes and failures without parameter updates, while efficiently navigating large frontier sets.  
We evaluate \explorer across multiple benchmarks including OpenEQA~\cite{majumdar2024openeqa} and GOAT-Bench~\cite{khanna2024goat} and find that it brings up to $3\times$ higher success rate and navigation efficiency than SOTA open-sourced baselines.


Our main contributions are as follows:
\begin{itemize}
  \item We introduce \explorer, a training-free framework that contextualize retrospective experience replay for embodied exploration, enabling non-parametric test-time scaling to optimize exploration decision-making.
  \item We propose a hierarchical frontier-ranking mechanism that efficiently handles large action spaces and enhances decision robustness.
  \item Our empirical study demonstrate substantial improvements in exploration efficiency and success across benchmarks (see~\cref{sec:experiments}), and provide ablations quantifying the impact of each module.
\end{itemize}
\section{Related Works}
\label{sec:relatedworks}

\subsection{MLLM-based Embodied Exploration}
MLLM-based embodied exploration~\cite{wang2025embodied} has recently accelerated through the integration of foundation models, semantic mapping~\cite{raychaudhuri2025semantic}, and memory-enhanced decision making~\cite{zhai2025memorycentricembodiedquestionanswer}. Early embodied QA and navigation systems relied on task-specific CNN–RNN~\cite{wang2016cnn} controllers trained via imitation or RL~\cite{zhou2025reinforced}, which struggled with sparse rewards and limited generalization~\cite{das2018embodied,wijmans2019embodied,yu2019multi}. 
ALFWorld~\cite{shridhar2021alfworld} innovatively uses a language-based policy on an exploration benchmark.
In the years to come, the emergence of large-scale pretrained models shifted this paradigm: multimodal models such as CLIP~\cite{radford2021learning}, BLIP-2~\cite{li2023blip}, GPT-4V~\cite{hurst2024gpt}, and MLLMs used in EfficientEQA~\cite{cheng2024efficient} now serve as in-the-loop semantic reasoners, enabling zero-shot high-level planning~\cite{hughes2025ai}, visual concept grounding~\cite{gu2024conceptgraphs}. Hierarchical designs, including Planner–Manager–Actor architectures in CityEQA~\cite{zhao2025cityeqa}, use LLMs for plan decomposition, semantic interpretation, and object-centric map construction~\cite{zhai2025memorycentricembodiedquestionanswer}, demonstrating improved interpretability and long-horizon reasoning~\cite{li2024memonavworkingmemorymodel}. Yet most early LLM-guided agents~\cite{booker2024embodiedrag} operated within single episodes and lacked mechanisms to accumulate knowledge over time~\cite{tan2023knowledge}. Recent advances address this by integrating persistent memory~\cite{li2024memonavworkingmemorymodel,zhai2025memory}, such as hierarchical episodic–semantic stores in MemoryEQA~\cite{zhai2025memory} and exploration histories in OpenEQA~\cite{majumdar2024openeqa}, allowing agents to recall visited regions, disambiguate landmarks, and reduce redundant exploration. Complementary frontier-driven frameworks~\cite{ren2024explore} prompt MLLMs with accumulated maps or semantic observations to identify informative frontiers~\cite{raychaudhuri2025semantic}, combining frontier reasoning with long-term memory to substantially improve exploration efficiency~\cite{cheng2024efficienteqa} and multi-task performance~\cite{yu2019multi}. Collectively, these developments mark a shift from geometry-only exploration toward semantically grounded~\cite{stranghoner2025share}, memory-augmented, MLLM-guided embodied agents~\cite{patel2024multillmqaembodiedexploration} capable of zero-shot generalization and cross-episode reasoning~\cite{kuehn2025humans}.

\subsection{Experience-inspired Embodied Agents}
Large Language Models have presented its excenllent text-test scaling potential~\cite{zhang2025survey}. This incubates a wide range of context engineering~\cite{mei2025survey} efforts to amplify the model knowledge at inference time. 
\cite{sarch2024vlm} learns abstract examples from human feedback as in-context input for programmable tasks.
\cite{liu2025contextual} presents a web agent that can can extract reusble skills for weg navigation tasks.
These initial attempts in programmable and verifiable tasks inspire us to extend the test-time experience reuse in more realistic embodied exploration field, where \emph{each step is hard to verify and action space is subtle}.
Noteworthily, we are \emph{not} emphasizing reusable memory/tool-use/programmable skill aquisition as in open-world gaming environment~\cite{zhai2025memory,lee2025enhancing}, where consolidated memory are strusctured and modular and have specific target.
we consider experience as a more abstrasct and generalizble that go beyond specific tasks and environment~\cite{lampinen2025latent}.
We differentiate it from episodic memory by (1) experience is concise and abstract; (2) experience is based on environment outcome.

\section{Methodology}
\label{sec:method}

\begin{figure*}
    \centering
    \includegraphics[width=1\linewidth]{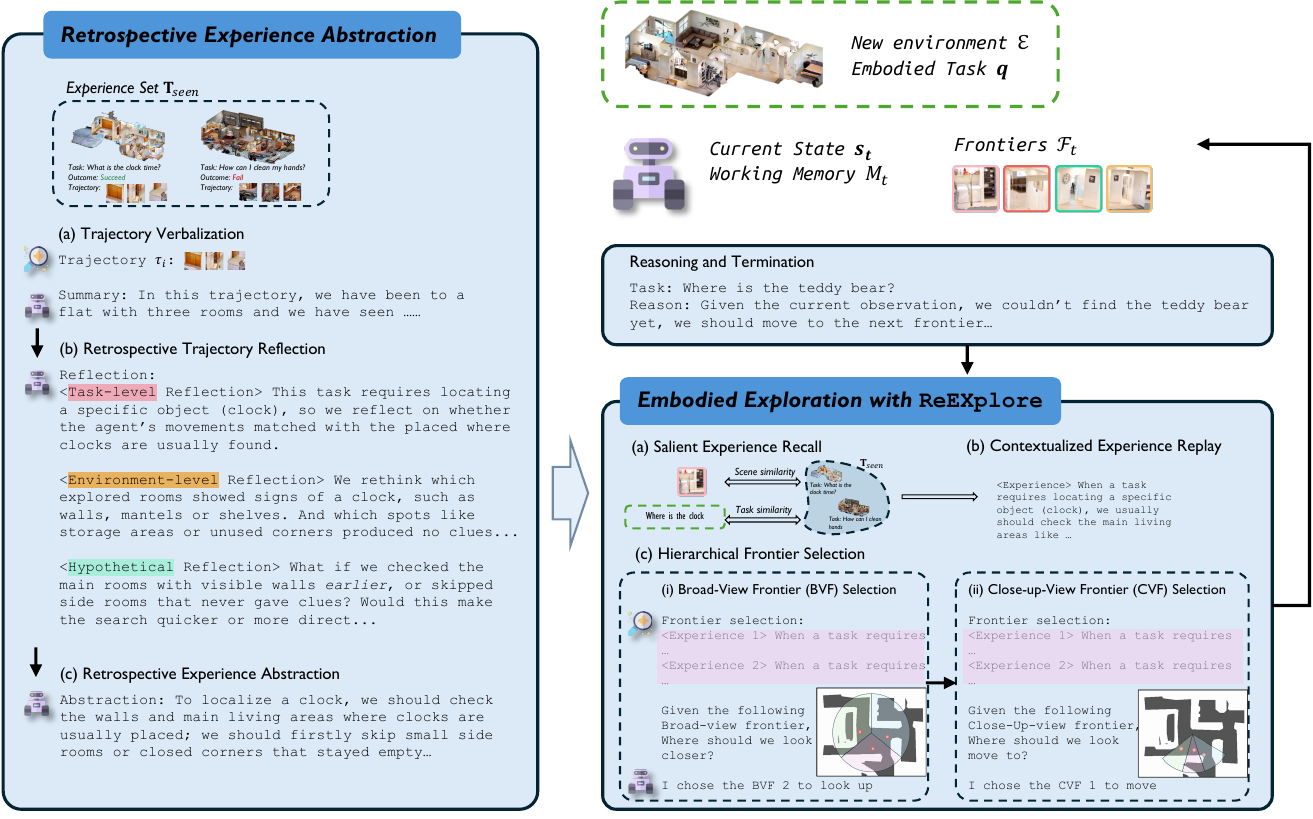}
    \caption{Overview of \explorer. \textbf{Left: Retrospective Experience Abstraction.}  
Completed trajectories are distilled into compact, transferable experience abstractions in a progressive way.  
\textbf{Right: Embodied Exploration with \explorer.}  
In a new environment, the agent retrieves salient past experiences based on scene and text similarity, incorporates them through contextualized experience replay, and performs hierarchical frontier selection to guide exploration.  This enables the agent to navigate toward informative viewpoints efficiently while leveraging distilled prior experience.}
    \label{fig:pipeline}
\end{figure*}

Our overall pipeline is illustrated in \cref{fig:pipeline}.
We consider an embodied agent tasked with exploring an unseen environment $\mathcal{E}$ to answer a question $q$. The agent starts at an initial location $l_0$ at step $t=0$ and continues exploring until it believes the target has been found or the maximum number of steps is reached.
At each step $t$, the agent’s state is defined as $\state = (\obs, \loc, \mem)$, where $\obs$ denotes the egocentric observation (\ie, RGB images), $\loc$ represents the agent’s spatial location, and $\mem$ corresponds to the working memory, a textual summary of the explored trajectory.
During exploration, given $\Ka$ Broad-View Frontier snapshots $\bvf$, which represent the unexplored boundaries of the current map, the agent retrieves a salient experience $\retr$ from the experience set $\expr$ and queries the policy model $\newpolicy$ to select a frontier $\selfron$ as the next action $a_t$. Within the selected region, referred to as Close-up-View Frontiers $\cvf$, the agent repeats the frontier selection process to determine a finer-grained direction for subsequent exploration.



\subsection{\explorer: Retrospective Experience For Embodied Exploration}
Embodied agents should learn from past for new situations.
In our work, we aim to break down the disction between \emph{episodic memory} and \emph{experience}.
Episodic memory traces from the subjective experience of remembering. Memory is stored information, while ``remembered experience'' experience as the distilled, structured knowledge abstracted across episodes~\cite{tulving1985memory,sutton1998reinforcement}.

This inspires us that instead of passively saving episodes/trajectory as raw events in a diverse exploration environment, it is more important to consolidate reusable experience as generalized schemas extracted across episodes~\cite{schank1999dynamic,gershman2010learning}. However, we can see so far experience-based learning is mostly a parametric way with reinforcement learning~\cite{zhu2023excalibur,zhou2025reinforced}, while training MLLMs for reinforcement learning requires (1) overheads for long-horizon tasks and (2) sparse reward for MLLMs.

Based on this, we propose a non-parametric and contextualization method, \explorer (\textbf{Re}trospective \textbf{EX}perience-based \textbf{EXplor}ation), consisting of two steps: Retrospective Experience Abstraction (See \cref{sec:rea}) and Contextualized Experience Replay (See \cref{sec:cer}).

\subsubsection{Retrospective Experience Abstraction}\label{sec:rea}
Experience highlights itself because we can see the outcomes of the trajectory and so we can reflect the trajectory based on the final outcome in a \emph{retrospective} way.
Given a completed trajectory $\tau$ and outcome $o$ in the training set, we abstract an experience via three steps.

\vspace{0.2cm}\noindent\textbf{Trajectory Verbalization.}
We convert each state $\state$ and the selected frontier $f_t^{(a_t)}$ in a trajectory $\tau$ into natural-language descriptions. To manage long-horizon trajectory and retain the local context of each step, we adopt a sequential chunking strategy: every 10 consecutive steps are grouped into one segment, whose step-level captions are summarized into a single chunk-level caption. 
All chunk captions are then aggregated into a coherent trajectory-level summary that captures the full exploration process for the given task. This hierarchical verbalization compresses long trajectories while preserving temporal structure and semantic continuity. We employ \qwen to generate both step-level and chunk-level captions.



\vspace{0.2cm}\noindent\textbf{Retrospective Trajectory Reflection.}
After each trajectory, we apply a self-reflection formatted prompt~\cite{shinn2023reflexion} to intuitively prompt the agent to explain why the exploration ultimately succeeded or failed, identifying which decisions were pivotal and what alternative choices could have produced a better outcome.  
We term this outcome-aware and globally focused reflection as \emph{retrospective experience}.

To ensure structured reasoning, we apply a standardized reflection procedure.  
The agent first restates the original question and specifies the success criterion.  
All explanations must rely strictly on observed regions, objects, and visual cues without fabricating unseen elements or referencing internal model states.  
This constraint is aligned with findings in previous work~\cite{shinn2023reflexion}, where disciplined textual feedback improves decision-making reliability.

We devise three key criterion for trajectory-level reflection:  
(1) \emph{task-level reflection}, decomposing the task itself and assessing whether the goal was achieved and which actions facilitated or hindered success;  
(2) \emph{environment-level reflection}, noting which regions, entities, or spatial cues in the current environment were informative or ambiguous; and  
(3) \emph{hypothetical reflection}, proposing plausible alternatives that could address mistakes or inefficiencies observed in the executed path or further improve the exploration.  

Beyond explaining past behavior, this reflection provides the analytical foundation for experience abstraction.

\vspace{0.2cm}\noindent\textbf{Retrospective Experience Abstraction.}
Experience represents structured, distilled knowledge rather than raw episodic memory.  
Thus, to enforce the trade-off between \emph{specificity} and \emph{generalizability}, we apply a chain-of-thought distillation process: the model re-traces its own reflection, identifies causal dependencies between cues, regions, and actions, and reformulates them into concise schema-like lessons.  
This CoT-based distillation removes trajectory-specific noise while preserving the causal mechanisms essential for transfer.  
Inspired by cognitive science and reflective agents~\cite{ge2025samule}, this structured two-paragraph abstraction yields robust, reusable experience that can guide future exploration without additional training.
The resulting experience encapsulates region-selection heuristics, evidence-driven pivots, semantic priors, and temporal considerations, forming a compact and reusable blueprint that informs future decision-making without relying on snapshot- or environment-specific details. The used prompt can be found in ~\cref{sup:abstraction_prompt}.

\subsubsection{Contextualized Experience Replay}\label{sec:cer}
During new episodes, we retrieve past experiences most relevant to the current state in order to guide frontier selection. We define the replay-augmented policy as:
\[
\newpolicy(a_t \mid \state,\fronts,\retr) 
= \Pr(a_t \mid \state,\fronts, \mathcal{R}_t),
\] 
where $\retr$ are recalled most salient retrospectice experience.

\vspace{0.2cm}\noindent\textbf{Salient Experience Recall.}
We begin by constructing an \emph{experience set} of pre-explored trajectories, each collected in a distinct environment.  For each question, the agent performs a full exploration episode, logging states and selected frontiers.  Every ten steps, consecutive actions are grouped into one \emph{chunk}, and \qwen summarizes each chunk into a \emph{chunk caption}.  All chunk captions are subsequently fused into one holistic \emph{trajectory-level caption}, which is then reflected upon to generate a two-paragraph \emph{trajectory abstraction}.  These abstractions form the experience library.

For each step $t$ in a new trajectory, the agent recalls past experiences through \emph{scene similarity} and \emph{task similarity}. 
(a) For each frontier snapshot in the candidates, we embed them with OpenCLIP ViT-H/14~\cite{Cherti_2023} and retrieve top-$m$ most similar scenes (frontier snapshots in the experience set $\mathcal{T}_{\text{seen}}$), which yields $K \times m$ matched stored snapshots. 
We then retrieve the trajectory index $\tau_i$ of the selected scenes and remove the duplicate trajectories, which retains the top-$n$ trajectories as \emph{visually relevant} experience to the current frontier selection. 
(b) For task similarity, we embed current question with a lightweight sentence embedding model (the MiniLM-L6-v2 variant from Sentence-Transformers\cite{reimers2019sentencebertsentenceembeddingsusing}), and using cosine similarity to compare them with all questions in the experience set $\mathcal{T}_{\text{seen}}$. Because each question corresponds to a trajectory, the top-$n$ most similar questions directly return the top-$n$ semantically related trajectories, as \emph{text relevant} experience to the current frontier selection.

We fuse both rankings using Reciprocal Rank Fusion (RRF) with constant $k{\approx}60$:
\[
\text{score}_{i} = \frac{1}{k + \text{rank}^{(\mathrm{scene})}_{i}}
                  + \frac{1}{k + \text{rank}^{(\mathrm{task})}_{i}}.
\]

RRF robustly prioritizes candidates that are jointly similar in visual appearance and semantic meaning.  The top-$K$ results are selected, and their associated trajectory abstractions constitute the replay context $\mathcal{R}_t$.  These retrieved experiences serve as distilled memories: summaries of strategies that previously succeeded or failed; and steer the agent toward more promising frontier choices. 

\vspace{0.2cm}\noindent\textbf{Working Memory.} We maintain a textual summary of the ongoing episode to support both reflection and real time reasoning. Working memory is a working memory of contextual buffer that summarizes the agent’s recent exploration within the same episode. 
At each step, the last five frontier snapshots are condensed into a brief paragraph describing visited regions, observed cues, and still-uncertain directions. 
This context supplies a continuous sense of spatial progress for frontier ranking, rather than treating observations as isolated frames. 
In contrast, a \textit{trajectory abstraction} is a cross-episode artifact: it is distilled from completed training episodes and then reused across different episodes as an experience prior. 
Thus, the working memory is \emph{episode-bounded and transient}, while the abstract experience is \emph{cross-episode and persistent}; together they provide complementary temporal scopes: short-horizon continuity versus long-term transferable knowledge.

\subsection{\textit{Divide-and-Conquer}: Hierarchical Frontier Selection}
\label{sec:hierarchical}


A key challenge in frontier-based exploration is the large action space: cluttered indoor scenes often expose dozens of frontiers at each step. Evaluating all of them with an MLLM is computation-heavys, sensitive to small viewpoint changes, and further affected by the model randomness. This makes direct ranking computationally costly and unreliable.

\vspace{0.2cm}\noindent\textbf{Hierarchical Frontier Partitioning.} To reduce this difficulty, we use a hierarchical selection with two layers, so the model can first choose a coarse direction and then make a more detailed choice as illustracted in~\cref{fig:pipeline}.

We first cluster the frontier set $\fronts$ with a two-stage K-means procedure. The first stage produces a fixed number of \emph{Broad-View Frontiers} (BVFs), each representing a coarse spatial region. In the second stage, every BVF is further divided into several \emph{Close-up-View Frontiers} (CVFs) that describe the finer areas inside that region. This creates a clear parent–child structure: each BVF contains its own group of CVFs.

The selection follows this hierarchy. The model first looks at the BVFs and chooses the region that best matches the question. After the BVF is selected, the model only considers the CVFs inside that BVF and picks one of them as the final frontier to explore. By limiting each decision to a small and fixed number of images, this two-layer process makes selection more efficient and stable, and it keeps spatially related frontiers in a consistent context even when the total number of candidates varies across steps. The used prompt can be found in ~\cref{sup:frontier_prompt}.

\subsection{Embodied Exploration Schema}
We integrate retrospective abstraction, and contextualized replay to form the complete exploration and task solving workflow, as show in ~\cref{fig:pipeline}. Further details can be found in ~\cref{sup:explore}.

\vspace{0.2cm}\noindent\textbf{MLLM Backbone.} 
We build on strong vision‑language models; in experiments we test an open‑source backbone (\eg, \qwen) and a commercial model (\eg, \gpt). 
Both models generate step-wise reasoning and action descriptions based on the current question, Working memory, and retrieved trajectory abstractions. 

\vspace{0.2cm}\noindent\textbf{Task Termination and Reasoning.} 
To determine the termination of the current task, we adopt the strategy in ~\cite{yang20253dmem3dscenememory} to finish the task.
This approach uses the \emph{memory snapshots} as inputs introduced in 3D-Mem when models prompted to answer the embodied question answering tasks. Memory snapshots are a rich representation of the Working memory of the current trajectory to help answer the question.
To note, we do not use memory snapshots for our embodied exploration.
At each step, the agent records several \textit{memory snapshots}, each containing a local RGB image and the set of detected object categories. 
Before reasoning, the model filters snapshots by matching question-relevant categories and keeps only those containing related objects. 
All remaining snapshots are presented to the model together with the question. 
If the model outputs an answer referring to a specific snapshot, that view is regarded as sufficient evidence; the agent moves to the same location where the snapshot was captured, produces the final answer, and terminates the task. 
Otherwise, exploration continues until such a view is found or no new areas remain. 
Snapshots are not used for navigation, since they only summarize past observations rather than actionable goals. 
They determine \textit{when} enough information has been gathered to answer, while frontier cues decide \textit{where} to explore next.

\section{Experimental Study}
\label{sec:experiments}

\begin{table*}[t]
\centering
\scriptsize
\adjustbox{max width=\textwidth}{
\begin{tabular}{lcccccccccccccccc}
\toprule
\textbf{Method} &
\multicolumn{2}{c}{\textbf{object recognition}} &
\multicolumn{2}{c}{\textbf{object localization}} &
\multicolumn{2}{c}{\textbf{attribute recognition}} &
\multicolumn{2}{c}{\textbf{spatial understanding}} &
\multicolumn{2}{c}{\textbf{object state recognition}} &
\multicolumn{2}{c}{\textbf{functional reasoning}} &
\multicolumn{2}{c}{\textbf{world knowledge}} &
\multicolumn{2}{c}{\textbf{overall}} \\

\midrule
\multicolumn{17}{l}{\textit{Socratic LLM-based Exploration w/ Frame Captions}} \\
GPT-4*   & 25.3 & --- & 28.4 & --- & 27.3 & --- & 37.7 & --- & 47.2 & --- & 54.2 & --- & 29.5 & --- & 35.5 & --- \\
\gpt   & 22.0 & --- & 25.0 & --- & 27.3 & --- & 40.8 & --- & 50.9 & --- & 61.8 & --- & 38.4 & --- & 35.9 & --- \\
\midrule
\multicolumn{17}{l}{\textit{Socratic LLM-based Exploration w/ Scene-Graph Captions}} \\
CG Scene-Graph Captions* & 25.3 & --- & 16.5 & --- & 29.2 & --- & 37.0 & --- & 52.2 & --- & 46.8 & --- & 37.8 & --- & 34.4 & 6.5 \\
SVM Scene-Graph Captions* & 29.0 & --- & 17.2 & --- & 31.5 & --- & 31.5 & --- & 54.2 & --- & 39.8 & --- & 38.9 & --- & 34.2 & 6.4 \\
LLaVA-1.5 Frame Captions* & 25.0 & --- & 24.0 & --- & 34.1 & --- & 34.4 & --- & 56.9 & --- & 53.5 & --- & 40.6 & --- & 38.1 & 7.0 \\
Multi-Frame* & 34.0 & --- & 34.3 & --- & 51.5 & --- & 39.5 & --- & 51.9 & --- & 45.6 & --- & 36.6 & --- & 41.8 & 7.5 \\
\midrule{\textit{Open-Sourced MLLM-based Exploration}} \\
3D-Mem (\qwen) & 25.0 & 13.9 & 23.8 & 7.7 & 52.6 & 29.3 & 33.3 & 3.9 & 57.5 & 26.1 & 43.8 & 13.5 & 37.5 & 8.4 & 39.1 & 14.6 \\
\textbf{\explorer (\qwen)} & \textbf{50.6} & \textbf{31.3} & \textbf{29.4} & \textbf{17.8} & 49.1 & \textbf{31.0} & \textbf{43.1} & \textbf{15.1} & \textbf{66.7} & \textbf{17.2} & \textbf{47.9} & \textbf{27.4} & \textbf{43.1} & \textbf{20.9} & \textbf{46.2} & \textbf{23.0} \\

\midrule{\textit{Commercial MLLM Exploration}} \\
Explore-EQA* (\gpt) & 44.0 & 19.6 & 37.1 & 29.6 & 55.3 & 36.0 & 42.1 & 6.6 & 46.3 & 9.2 & 63.2 & 35.7 & \textbf{45.5} & 22.0 & 46.9 & 23.4 \\
CG w/ Frontier Snapshots* (\gpt) & \textbf{45.0} & \textbf{42.0} & 32.1 & 25.0 & 50.8 & 35.2 & 32.9 & 18.7 & 68.5 & 38.4 & 58.8 & 42.2 & \textbf{45.5} & \textbf{33.5} & 47.2 & 33.3 \\
3D-Mem (\gpt) & 35.0 & 18.8 & 50.0 & 37.3 & 64.3 & \textbf{56.3} & 50.0 & 24.7 & 80.0 & 49.3 & 50.0 & 22.1 & 30.0 & 21.4 & 54.4 & 33.3 \\ 
\textbf{\explorer (\gpt)} & 37.5 & 21.9 & \textbf{65.6} & \textbf{45.8} & \textbf{67.9} & 46.1 & \textbf{60.0} & \textbf{25.3} & \textbf{100.0} & \textbf{53.5} & \textbf{65.0} & \textbf{42.4} & 35.0 & 23.6 & \textbf{58.3} & \textbf{37.3} \\

\midrule
Human Agent* & 89.7 & --- & 72.8 & --- & 85.4 & --- & 84.8 & --- & 97.8 & --- & 78.9 & --- & 88.5 & --- & 85.1 & --- \\

\bottomrule
\end{tabular}
}
\caption{Performance on A-EQA by Question Categories. For each question categories, there are two columns. The first column stands for the \llm Score, while the second column represents the \llmspl score. “CG” denotes ConceptGraphs. Methods with * are reported from OpenEQA\cite{majumdar2024openeqa} and 3D-Mem\cite{yang20253d}. Columns represent different category of questions in the dataset.}
\label{tab:llm}
\end{table*}

\subsection{Active Embodied Question Answering}
We evaluated our method on the embodied question answering benchmark, Open-EQA~\cite{majumdar2024openeqa} for Active EQA (A-EQA) to test exploration efficiency. 
On the A-EQA benchmark (\cref{tab:llm} and \cref{tab:succ_spl}), we evaluate our framework \explorer along with other MLLM-based baselines  designed to enable embodied exploration over complex open-ended questions. 

\vspace{0.2cm}
\noindent\textbf{Experience Set.}  
A-EQA is an open-vocabulary benchmark built upon HM3D~\cite{ramakrishnan2021habitat} scenes, covering diverse reasoning skills such as object recognition, spatial understanding, and functional reasoning. 
We construct the set of experiences with environments and tasks from the A-EQA training split, which has 52 different environments and 164 tasks to generate retrospective experience.

\vspace{0.2cm}
\noindent\textbf{Metrics.}  
We first report the Success Rate (Succ.) and SPL, which measures navigation efficiency by comparing the agent’s path length with the shortest possible path (0–1 scale). The higher SPL is, the closer the agent’s trajectory is to the optimal one.

We then follow the OpenEQA\cite{majumdar2024openeqa} protocol to report \textit{\llm} and \textit{\llmspl}. \llm evaluates the semantic correctness of the model answers using \gpt, and \llmspl further weights this score by normalized path length. If the agent does not respond, \gpt is queried without visual input and the SPL term is set to zero.

\vspace{0.2cm}
\noindent\textbf{Analysis.}
As shown in \cref{tab:llm}, \explorer achieves clear and consistent gains over all baselines under both \qwen and \gpt backbones.
With \qwen, overall \llm improves from 39.1 to 46.2, and \llmspl increases from 14.6 to 23.0.
Most task categories exhibit substantial improvements: object recognition rises from 13.9 to 31.3, object localization from 7.7 to 17.8, spatial understanding from 3.9 to 15.1, and functional reasoning from 13.5 to 27.4.
Since \llm evaluates the accuracy and plausibility of the final prediction, higher scores imply that the agent concludes its trajectory at more informative and semantically meaningful viewpoints.
The strong \llm gains therefore indicate that \explorer consistently guides the agent toward evidence-rich locations earlier in the episode, instead of relying on incidental late-stage observations.
This finding is reinforced by the trends in \cref{tab:succ_spl}, where both success rate and SPL improve from 50.9\% and 36.4\% to 58.5\% and 52.9\%, respectively, showing that performance is not compromised while exploration paths significantly are optimized.

With \gpt, the trend remains consistent: \explorer reaches 58.3 in \llm and 37.3 in \llmspl (vs.\ 54.4/33.3 for 3D-Mem). 
Category-specific gains are similarly strong—for instance, \llm reaches 100.0 with 49.3 \llmspl on object-state questions, and SPL improves by 7–10 points on attribute and localization tasks without reducing success.
From \cref{tab:succ_spl}, the overall Success rate and SPL rise from 69.6\%/61.9 to 75.5\%/65.2, confirming that \explorer consistently enables faster and more efficient acquisition of crucial evidence.

Across categories, \explorer yields consistently higher \llm (indicating better final answer quality) and higher \llmspl/SPL (indicating more efficient trajectories), showing that the agent not only identifies the correct evidence but also reaches it through more direct and purposeful exploration.
The improvements are particularly pronounced in \textit{Object Recognition}, \textit{Object Localization}, and \textit{Object State Recognition}, suggesting that retrospective experience is especially beneficial for object-centric tasks where recognizing or locating specific entities is crucial.

Taken together, \explorer improves \llm, reflecting more accurate and better-grounded answers; while \llmspl/SPL, reflecting more efficient movement toward informative viewpoints—across both model families.
The joint improvements provide convergent evidence that \explorer consistently guides the agent to the right places and guides it there more efficiently.

\begin{figure}
    \centering
    \includegraphics[width=1\linewidth]{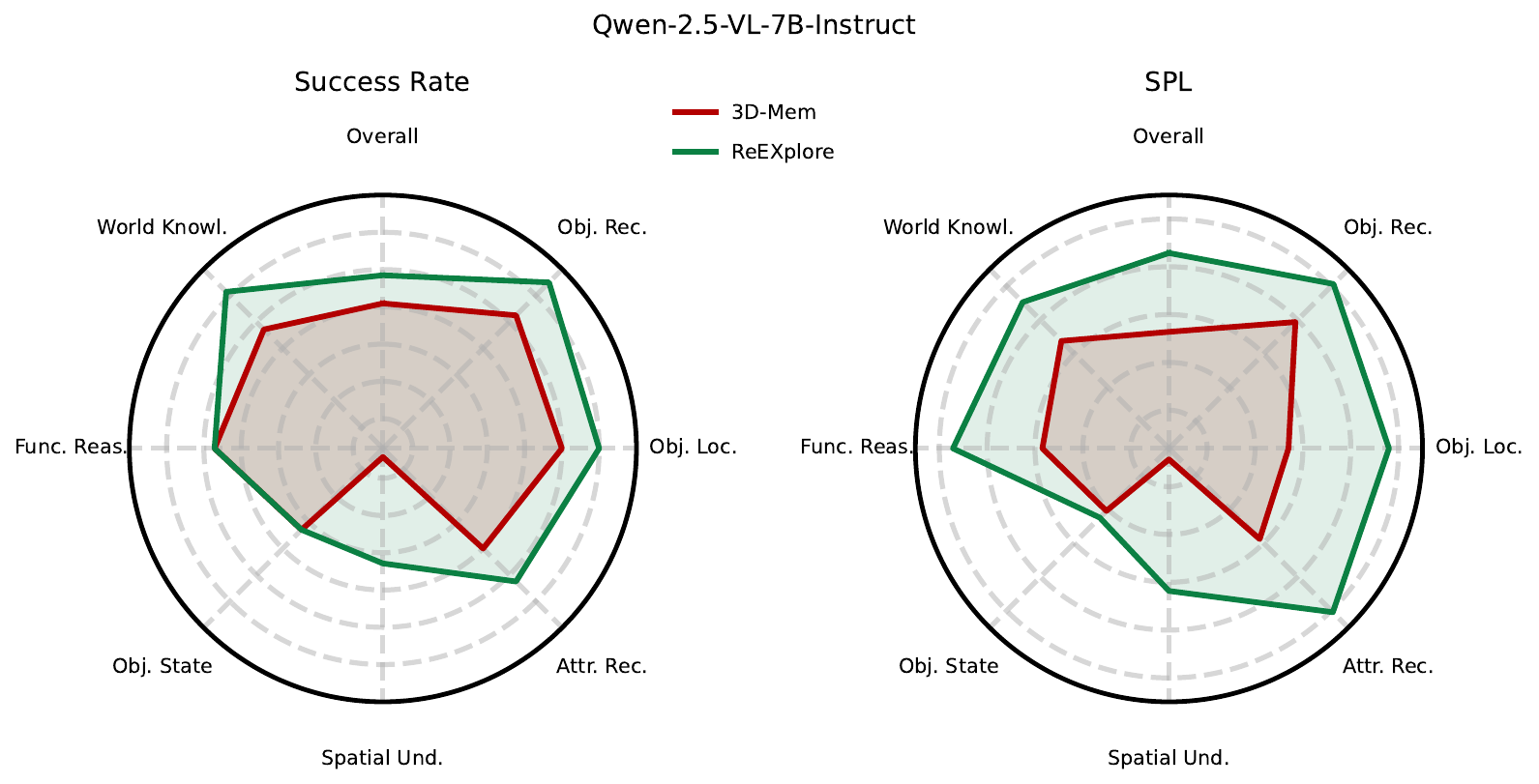}
    \caption{Performance of \explorer (with \qwen) on A-EQA by question categories, showing both \textbf{Success Rate (Succ.)} and \textbf{SPL}. Our approach outperms the strongest baseline model by a large margin in all task categories.}
    \label{fig:qwen-succ_spl}
\end{figure}


\begin{table}[ht!]
\centering
\scriptsize
\begin{tabular}{lcc}
\toprule
\textbf{Method} & \textbf{Success Rate ($\%$)} $\uparrow$ & \textbf{SPL ($\%$)} $\uparrow$ \\
\midrule
\multicolumn{3}{l}{\textit{\textbf{Traditional Methods}}} \\
Modular GOAT~\cite{DBLP:conf/rss/Chang_goat}*                 & 24.9 & 17.2 \\
Modular CLIP on Wheels~\cite{gadre2023cows}*       & 16.1 & 10.4 \\
SenseAct-NN Skill Chain~\cite{khanna2024goat}*      & 29.5 & 11.3 \\
SenseAct-NN Monolithic~\cite{khanna2024goat}*       & 12.3 &  6.8 \\
\midrule
\multicolumn{3}{l}{\textit{\textbf{MLLM-based Exploration}}} \\
3D-Mem~\cite{yang20253dmem3dscenememory}$^\dagger$ (\qwen) & 49.4 & 20.7 \\
\explorer$^\dagger$ (\qwen) & \textbf{53.2} & \textbf{32.6} \\
\cmidrule{1-1}
Explore-EQA~\cite{ren2024explore}$^\dagger$ (\gpt)                & 55.0 & 37.9 \\
CG w/ Frontier Snapshots~\cite{gu2024conceptgraphs}$^\dagger$ (\gpt)   & 61.5 & 45.3 \\
3D-Mem w/o memory~\cite{yang20253dmem3dscenememory}$^\dagger$  (\gpt)         & 58.6 & 38.5 \\
3D-Mem~\cite{yang20253dmem3dscenememory}$^\dagger$(\gpt) & \textbf{65.1} & \textbf{49.3} \\
\explorer$^\dagger$(\gpt) & 59.8 & 42.5 \\
\bottomrule
\end{tabular}
\caption{\textbf{Experiments on GOAT-Bench} on the ``Val Unseen" split. ``CG" denotes ConceptGraphs. Methods denoted by * are taken from GOAT-Bench, and those marked with $^\dagger$ are evaluated on the subset specified in ~\cite{yang20253dmem3dscenememory}.}
\label{tab:goatbench}
\end{table}

\begin{table*}[t]
\centering
\scriptsize
\adjustbox{max width=\textwidth}{
\begin{tabular}{lccccccc}
\toprule
\multirow{2}{*}{\textbf{Method}} 
& \multirow{2}{*}{\begin{tabular}{c}
\textbf{Retrospective} \\
\textbf{Experience Replay}
\end{tabular}}
& \multirow{2}{*}{\begin{tabular}{c}
\textbf{Working} \\
\textbf{Memory}
\end{tabular}}
& \multirow{2}{*}{\begin{tabular}{c}
\textbf{Hierarchical} \\
\textbf{Frontier Selection}
\end{tabular}}
& \multicolumn{4}{c}{\textbf{Overall}} \\

\cmidrule(lr){5-8}
 &  &  &  & \textbf{LLM-Match} & \textbf{LLM-Match$\times$SPL} & \textbf{SPL} & \textbf{Succ.} \\
\midrule

\explorer (\qwen) & \checkmark & \checkmark & \checkmark & \textbf{46.2} & \textbf{23.0} & \textbf{52.9} & \textbf{58.5} \\
 &  & \checkmark & \checkmark & 40.3 & 20.0 & 48.6 & 52.6 \\
 &  &  & \checkmark & 37.5 & 15.9 & 36.9 & 49.5 \\
 &  &  &  & 34.2 & 15.9 & 45.1 & 56.2 \\

\midrule

\explorer (\gpt) & \checkmark & \checkmark & \checkmark & \textbf{58.3}  & \textbf{37.3} & \textbf{65.2} & \textbf{75.5} \\
 &  & \checkmark & \checkmark & 50.6 & 31.6 & 63.0 & 73.7 \\
 &  &  & \checkmark & 42.3 & 25.7 & 61.9 & 71.6 \\
 &  &  &  & 50.7 & 32.4 & 61.3 & 66.1 \\

\bottomrule
\end{tabular}
}
\caption{Ablation studies by ablating the key components in \explorer (Retrospective Experience Replay, Working Memory, and Hierarchical Frontier Selection) with two backbone MLLMs.}
\label{tab:ablation}
\end{table*}

\subsection{Lifelong Visual Navigation}
GOAT-Bench~\cite{khanna2024goat} is a lifelong visual navigation benchmark featuring a multi-target Go-to-AnyThing (GOAT) task. An agent needs to sequentially negative to different objects in an unseen environments. The object are specified by either a category name (\eg, refrigerator), a language description (\eg, dresser next to the armchair), or an image of the target.  Under such a long trajectory scenario, abstract experience and hierarchical frontier selection should be of vital importance to navigate in a more accurate and efficient manner. To keep consistent with our most important baseline method ~\cite{yang20253dmem3dscenememory}, we also evaluate \explorer on a 1/10-size subset of the ``val-unseen'' split, with one episode on each of the 36 scenes, totaling 278 navigation subtasks. 

\vspace{0.2cm}\noindent\textbf{Experience Set.}
We use the ``train'' split from GOAT-Bench and run rollouts with hierarchical frontier selection described in \cref{sec:hierarchical} without any experience replay. In summary, we collect 255 trajectories in 34 diverse scenes, a similar size as the evaluation set, incorporating both successful and failure trials. The experience is generated following \cref{sec:rea} and injected during evaluation following \cref{sec:cer}.

\vspace{0.2cm}\noindent\textbf{Metrics.} We report the Success Rate and Success weighted by Path Length (SPL). A successful navigation task means that the agent is located within 1 meter from the navigation goal. SPL is the success weighted by the trajectory length.

\vspace{0.2cm}\noindent\textbf{Analysis.} 
As shown in \cref{tab:goatbench}, \explorer outperforms most baseline methods, demonstrating a clear shift toward more decisive and efficient exploration, as reflected by higher success rates and SPL scores. In particular, when compared to 3D-Mem with the Qwen backbone, the SPL improvement from 20.7\% to 32.6\% is substantially larger than the corresponding increase in success rate. This suggests that the agent achieves success not through extensive coverage but by committing earlier to more informative viewpoints.
In ~\cref{tab:goatbench-full} in Supplementary Material, the proposed experience abstraction notably improves success rates for the object and image categories, while offering limited gains for language descriptions, likely due to linguistic complexity and environmental diversity, which hinder relevant experience retrieval. Nevertheless, SPL improves consistently across all categories, confirming that experience-based recall effectively reduces redundant exploration.
In the \gpt setting, our method slightly trails 3D-Mem, mainly because prompt tuning was not feasible due to budget constraints. Further details are provided in ~\cref{sup:goatbench_result}. Overall, the results show a clear transition from wide searching to focused, purposeful movement, yielding higher success rate through shorter trajectories.


\subsection{Ablation Study}

\cref{tab:ablation} reports ablations on both the \qwen and \gpt backbones. 
We start from the full \explorer agent, which leverages  \textit{Retrospective Experience Replay}, short-term \textit{Working Memory}, hierarchical frontier selection and then progressively disable modules (\textit{w/o} means \textit{without}). 
Removing the Experience module (\textbf{w/o Retrospective Experience Replay}) already leads to clear drops in accuracy and efficiency 
(\ie, \llm $58.3\rightarrow50.6$ and \llmspl $37.3\rightarrow31.6$ with \gpt; 
$46.2\rightarrow40.3$ and $23.0\rightarrow20.0$ on \qwen), 
showing that long-term trajectory abstractions provide useful transferable priors for decision making. 
Further removing the Working Memory (\textbf{w/o Retrospective Experience/Working Memory}) leaves only the hierarchical frontier selection and causes additional degradation, especially in SPL and success 
(\eg, \qwen success rate 52.6$\rightarrow$49.5; GPT-4o 73.7$\rightarrow$71.6), 
indicating that maintaining a short-term textual context helps avoid redundant backtracking and improves navigation efficiency.
The last variant ablates all three components.
We notice that for both backbone MLLMs, the performance drops non-marginally in all dimensions. This again validates the synergy of the all three components. Further details can be found in ~\cref{sup:qual_analysis}.



\section{Conclusion}
\label{sec:conclusion}
We introduced \explorer, a training-free framework that enhances MLLM-based embodied agents through retrospective experience abstraction and contextualized experience replay. Our approach reframes embodied exploration as a non-parametric policy adaptation problem, enabling agents to incorporate distilled experiences from past trajectories directly at inference time. Complemented by a hierarchical frontier selection mechanism, \explorer supports reliable and efficient exploration in large and visually nuanced action spaces.  
Extensive experiments across multiple benchmarks demonstrate that \explorer consistently improves both embodied question answering accuracy and exploration path efficiency over strong MLLM baselines, regardless of model scale or task category. These results highlight the value of structured, reusable experience for guiding exploration at inference time.

\vspace{0.2cm}
\noindent\textbf{Limitations}
While \explorer improves MLLM-based exploration in indoor environments, its effectiveness in more diverse and open-ended outdoor settings remains underexplored. Outdoor scenes exhibit far greater visual and task variation, where experience should play an even more critical role. Extending our retrospective experience replay to such settings is an important direction for future work.

{
    \small
    \bibliographystyle{ieeenat_fullname}
    \bibliography{main}
}

\clearpage
\setcounter{page}{1}
\appendix
\newpage

\section*{Appendix}

This supplementary material provides additional technical details, extended experiments, and qualitative analyses that complement the main paper. 
~\cref{sup:implementation} describes our full implementation pipeline, including the setup, the implementation of \explorer (retrospective experience abstraction, hierarchical frontier selection, embodied exploration pipeline), and benchmark-specific configurations. 
~\cref{sup:extended} presents extended quantitative results on both A-EQA\cite{majumdar2024openeqa} and GOAT-Bench\cite{yang20253d}, including hyperparameter settings, significance analyses, and comparisons across different frontier-selection baseline methods with open-sourced models. 
Finally, ~\cref{sup:qual_analysis} provides qualitative analyses that illustrate how retrospective experience replay and hierarchical frontier selection impact exploration behavior and efficiency.

\section{Implementation Details}
\label{sup:implementation}

This section elaborates the  implementation details of \explorer. 
We begin by outlinining the embodied exploration simulation environment that we are based on and evaluation protocol in ~\cref{sec:platform}. 
We then detail the implementation of  retrospective experience abstraction, hierarchical frontier-selection, and their integration into embodied exploration in ~\cref{sup:rea_implementation,sup:hier,sup:explore}. 
Finally, we present the benchmark configurations of our experiments in \cref{sec:benchmarks}.

\subsection{Platform \& Implementation}
\label{sec:platform}
\vspace{0.2cm}\noindent\textbf{Platform \& Implementation.}
Our framework is implemented based on the codebase of Explore-EQA~\cite{ren2024explore} and 3D-Mem~\cite{yang20253dmem3dscenememory}. All experiments are conducted in Habitat-Sim~\cite{habitat19iccv,szot2021habitat,puig2023habitat3} using HM3D scenes. At each step, the agent receives RGB images if the environment and all actions are executed via a shortest-path planner.

For reasoning and answering questions, we follow 3D-Mem~\cite{yang20253d} to maintain a compact record of explored regions. Each RGB frame is first processed by a 2D object detector, and the detected objects are lifted into 3D using the corresponding depth map to form a unified global object set~\cite{roger2025robin}. Co-visible objects are then grouped through a co-visibility clustering procedure, where the RGB-D information is used only to estimate 3D proximity and cluster membership~\cite{liang2024centered}. After clustering, we retain a single representative \emph{RGB} frame as the memory snapshot for each cluster, capturing several co-visible objects along with their local visual context~\cite{liu2025contextual}. \emph{These memory snapshots are used only for question answering and are not used for exploration in \explorer.}

For exploration, we maintain an occupancy map that marks explored and unexplored areas. The boundary between these regions defines a set of frontiers, and the agent records a \emph{frontier snapshot} for each frontier by imaging the corresponding direction~\cite{yang20253dmem3dscenememory}.


\vspace{0.2cm} \noindent \textbf{Metrics.}
The metrics are calculated as follows:
For each question \(q_i \in Q\) (with \(|Q|=N\)), we record:  
(1) the oracle shortest-path distance \(G_i\),  
(2) the executed path length \(P_i\) (set to \(+\infty\) if no answer is produced),  
(3) the predicted answer \(a_i\),  
(4) an LLM-based correctness score \(s_i\in[1,5]\) when available, and  
(5) a baseline score \(b_i\in[1,5]\) used only as a fallback.  
A lightweight format check determines whether \(a_i\) is syntactically valid; we denote the set of valid answers by  
\[
S=\{\,q_i\in Q\mid a_i\ \text{is valid}\,\}.
\]

\noindent
\textbf{\textit{Success Rate.}} quantifies the proportion of questions for which the system produces a valid answer. Formally, let $N$ denote the total number of questions and $S$ the subset of questions with valid outputs. The metric is defined as:
\[
\mathrm{Success \;Rate}=\frac{|S|}{N}.
\]
In practice, we report the final score as a percentage by multiplying the computed ratio by 100.

\vspace{0.2cm}
\noindent\textbf{\textit{Path Efficiency (SPL).}}
We evaluate navigation efficiency using the standard SPL metric. For each question $q_i$, let $G_i$ denote the shortest oracle path length and $P_i$ the executed path length. The per-question score is:
\[
\mathrm{SPL}_i=\frac{G_i}{\max(G_i,P_i)},
\qquad
\mathrm{SPL}=\frac{1}{N}\sum_{q_i\in Q}\mathrm{SPL}_i.
\]
$\mathrm{SPL}_i$ measures how close the agent's trajectory is to the optimal route: it equals $1$ when the agent follows the shortest possible path, and decreases as the executed path becomes longer or detours occur. When the system fails to answer, we set $\mathrm{SPL}_i=0$.

In practice, we report the final SPL score in percentage form by multiplying the computed value by 100.

\vspace{0.2cm}
\noindent\textit{\textbf{\llm.}}
\llm as the semantic correctness of predicted answers is assessed with an external LLM-based judge~\cite{majumdar2024openeqa}. Given the question, the ground-truth answer, and the model's prediction, the grader assigns a discrete score $x\in\{1,2,3,4,5\}$ that reflects how well the prediction captures the intended meaning. A score of $5$ indicates a fully correct and semantically aligned answer, while $1$ corresponds to an unrelated or clearly incorrect response. The grading prompt contains explicit scoring rules and several illustrative examples, enabling consistent and human-like judgments.

When multiple paraphrases of the ground-truth answer are known, they are also provided to the grader so that semantically equivalent predictions can receive high scores even if their surface forms differ.
If the system fails to produce any valid answer, the grader will not be used and the item will be assigned a score of $0$, marking a ``no response'' case rather than a low-quality answer.

All LLM scores on the 1-5 scale are converted to a 0-100 range via a linear mapping:
\[
\varphi(s_i)=100\cdot\frac{s_i-1}{4}.
\]
Only questions that both (1) receive a valid system output and (2) have an LLM grader score contribute to the metric. Let
\[
P=\{\,q_i\in Q \mid s_i\ \text{exists}\,\},
\]
the final \llm score is then computed by averaging the mapped scores across all eligible questions:
\[
\text{\llm}
=\frac{1}{|S\cap P|}
\sum_{i\in S\cap P}\varphi(s_i).
\]
This produces a single 0-100 semantic accuracy score representing the overall mean quality of the system's valid and reliably graded answers.

\vspace{0.2cm}
\noindent\textit{\textbf{\llmspl.}}
This metric jointly evaluates two factors of embodied QA performance:  
(1) the semantic correctness of the final answer, and  
(2) the navigation efficiency of the executed trajectory.  
Both components are necessary for a question to be considered well-solved.
For each question $q_i$, we first derive a semantic score. When an LLM-based grader provides a score $s_i\in\{1,\dots,5\}$, we use it directly. If no grader score exists (e.g., the grader was not triggered or the answer was unsuitable for grading), we use a baseline fallback score $b_i$. This fallback score is a conservative estimate derived from a lightweight rule-based format check, ensuring that questions without a grader score still obtain a meaningful but non-inflated semantic value:
\[
\hat{s}_i=
\begin{cases}
s_i, & s_i\ \text{exists},\\
b_i, & \text{otherwise}.
\end{cases}
\]
We then map each $\hat{s}_i$ from the 1-5 scale onto a 0-100 range:
\[
\varphi(\hat{s}_i)=100\cdot\frac{\hat{s}_i-1}{4}.
\]
This produces a normalized semantic-correctness score where higher values indicate closer alignment with the ground truth.
Each question also has a navigation-efficiency score $\mathrm{SPL}_i$, which is near $1$ when the agent follows an almost optimal path, and near $0$ when the path is inefficient or the target is not reached.

To ensure that strong performance on only one dimension does not dominate the evaluation, we compute a joint score for each question by multiplying the semantic and navigation components. The final metric averages this joint score across all questions:
\[
\text{\llm}\times\mathrm{SPL}
=\frac{1}{N}\sum_{q_i\in Q}\left(\varphi(\hat{s}_i)\cdot\mathrm{SPL}_i\right).
\]
The result is a unified 0-100 metric that rewards agents only when they both answer correctly and navigate efficiently.

Success Rate reflects the agent’s reliability in producing valid answers; SPL quantifies its exploration efficiency; \llm measures semantic correctness; and the combined \llm\(\times\)SPL captures the overall effectiveness of reasoning and navigation jointly. These four metrics provide a balanced and comprehensive evaluation of navigation performance.

\vspace{0.2cm} \noindent \textbf{Evaluation Protocol.}
Each episode allows at most 50 interactive exploration steps as computation/time budgets. MLLMs share a unified configuration (including exploration and retrospective experience abstraction): a sampling temperature of 0.7, a maximum output length of 4096 tokens, and top-$p$ truncation at 0.95. These inference-time settings are identical for all methods evaluated.

All experiments operate under a unified sensing and perception setup. The onboard camera is mounted at a height of 1.5\,m with a downward pitch of $-30^\circ$ and a horizontal FOV of $120^\circ$. Images are rendered at $1280\times1280$ and uniformly resized to $360\times360$ before being provided to the model. For 3D mapping, we maintain a dense TSDF volume with a voxel size of 0.1\,m and fuse depth only within a 1.7\,m effective range. When updating the occupancy grid, the upper and lower 60\% of the image and the left and right 25\% are masked out to enforce conservative safety margins. Semantic perception adopts a deliberately low detection-confidence threshold (0.003) to maximize recall, followed by 2D non-maximum suppression with a 0.1 IoU threshold. In 3D, only detections within 3.5\,m of the camera are retained, and an IoU of at least 0.6 is required to associate a detection with the queried semantic target. All sensing, mapping, and perception parameters remain fixed across all evaluated methods.

\subsection{Implementation of Retrospective Experience Abstraction}
\label{sup:rea_implementation}
We use the training split of both datasets for retrospective experience abstraction.
Each episode ends with a complete trajectory $\tau$ and a final exploration outcome $o\!\in\!\{\text{PASS},\text{FAIL}\}$. Since the entire trajectory is fully observed only after the episode terminates, we can analyze it in a \emph{retrospective} manner by looking back at where the agent moved, what information it gathered, and how these choices relate to the final result. Our offline pipeline distills each trajectory into reusable experience through two stages: (1) trajectory verbalization and (2) trajectory-level reflection and abstraction. This section provides the implementation details that complement the main text in~\cref{sec:rea}.

\vspace{0.2cm}
\noindent\textbf{Trajectory Verbalization.}
For every trajectory in the training split, we reconstruct a chronological sequence of step tuples
\[
\{(t,\;\text{text}_t,\;f_t)\}_{t=0}^{T-1},
\]
where $\text{text}_t$ is a short natural-language description logged during exploration and $f_t$ indicates the chosen frontier direction at step $t$. These directional indicators are used only as implicit cues for later summarization and never appear in the final output.

Since a full trajectory may contain dozens of steps, we adopt a chunk-wise summarization scheme. Consecutive steps are grouped into fixed-length segments of 10. Each segment is passed to an LLM that receives: (i) the target question, (ii) the final outcome $o$, and (iii) the segment’s ordered list of step descriptions. For each segment, the model distills it into a concise single graph, which shows how agent move in this temporal segment as a chunk caption.
All chunk captions will be then merged and further consolidated, producing a single, coherent, and objective description of the full exploration trajectory. This summary highlights the sequence of visited regions, the structural layout of the environment, and the major directional transitions, without restating the question text, the success/failure outcome, or any implementation details. The resulting text is treated as the trajectory caption for next stage.

\vspace{0.2cm}
\noindent\textbf{Trajectory-Level Reflection and Abstraction.}
The reflection prompt is organized around the three levels of reflection illustrated in \cref{fig:pipeline}. (1) \emph{task-level} reflection encourages the model to restate the goal and evaluate whether the agent’s movements were consistent with what the question required. This corresponds to the task understanding and the high-level trajectory recounting in the early parts of the prompt. (2)  \emph{environment-level} reflection prompts the model to examine how the episode interacted with the scene layout, including which rooms or connectors offered meaningful evidence, which areas consistently produced none, and how typical object-region relationships shaped the search. These considerations align with the prompt elements that summarize spatial progression and capture general region-object associations. (3)  \emph{hypothetical} reflection prompts the model to reflect on alternative choices, for example giving earlier priority to more promising regions or skipping areas that repeatedly yielded no clues, and to identify behaviors that could lead to inefficiency or failure. This perspective aligns with the prompt components that focus on directional reasoning and patterns of ineffective behavior. The final abstraction integrates these three perspectives into a transferable and scene-agnostic exploration strategy that can serve as retrospective experience for future episodes. The detailed prompt have been shown in \cref{sup:abstraction_prompt}.

\subsection{Implementation of Hierarchical Frontier Selection}
\label{sup:hier}

A core difficulty in frontier-based exploration is that cluttered indoor scenes often expose a large number of potential frontiers. Directly ranking all of them with an MLLM is expensive and unstable, since small viewpoint changes can alter the model's judgment. With hierarchical frontier selection, we decompose the long list of candidate frontiers into an interpretable hierarchy that enables the model first choose a broad direction and then refine its choice to a specific viewpoint.

\vspace{0.2cm} \noindent \textbf{Binary maps and Frontier Formation.}
The agent observes the environment using a camera that provides, for every part of the image, an estimate of how far that part of the scene is from the robot.  
To turn these observations into a spatial representation, we place a regular 3D grid of small cubes (voxels) around the robot, each cube representing a $0.1\,$m region in the real world.  
Whenever the camera sees a piece of the scene, we update the corresponding cubes in the grid: cubes that appear to contain open space are marked as empty, cubes that appear to contain a surface are marked as occupied, and all cubes touched by the observation are marked as “seen at least once.”  
Over time, as the robot collects more observations, this grid becomes a coarse 3D sketch of the parts of the environment that have actually been seen.

To use this information for navigation, we convert the 3D grid into a 2D map.  
Since the robot moves on the floor, we only need to know, for each ground location, whether it is free to walk on, blocked by an object, or simply not yet observed.  
For every horizontal position $(x,y)$, we examine the vertical stack of cubes above it:  
if the stack contains clear space around the robot’s height and solid support below, we classify the location as free;  
if the stack has never been observed, it is marked unexplored;  
and if the stack repeatedly shows an object around the robot’s height, the location is marked occupied.  
This produces a top-down 2D grid where each cell summarizes the local geometry relevant for movement.

We work on this 2D grid and derive three binary layers used for exploration:  
(i) \textit{unoccupied}, the set of free and physically supported cells;  
(ii) \textit{unexplored}, the set of cells whose voxel stacks have never been observed;  
(iii) \textit{island}, the connected component of unoccupied cells that contains the agent, representing the region that is currently reachable.

To identify boundary cells between known and unknown space, we compute for each reachable grid cell $\mathbf{p}$ a frontier score using a \(3\times3\) convolution over the unexplored mask:
\[
s(\mathbf{p}) = \sum_{\mathbf{v} \in \mathcal{N}(\mathbf{p})} \mathbf{1}_{\text{unexplored}}(\mathbf{v}),
\]
where $\mathbf{v} \in \mathcal{N}(\mathbf{p})$ are neighboring grid cells. A grid cell is classified as a frontier when it lies inside the reachable island and has a score within a valid range:
\[
\text{frontier}(\mathbf{p}) =
\mathbf{1} \bigl[\, 
\mathbf{p} \in \textit{island}\
\land\
\tau_{\min} \le s(\mathbf{p}) \le \tau_{\max} 
\,\bigr].
\]
This produces a thick frontier band marking the transition from explored space to unseen regions. A stricter threshold yields a thinner “edge” band that we later use to anchor frontier centers. If either band is empty, the step provides no valid frontiers.

\vspace{0.2cm} \noindent \textbf{Broad-View Frontier (BVF) Construction.}
Given the reachable frontier set $\fronts=\{\mathbf{p}_i\}$ extracted from the occupancy maps, we aim to group these frontier points into a small number of coarse exploration directions. Each frontier point $\mathbf{p}_i=(p_i^x,p_i^y)$ is a 2D location on the robot-centric ground plane and represents a position along the boundary between explored and unexplored space. Clustering these 2D points helps identify where the frontier band is concentrated in different directions.
We set\[k=\min(3,|\fronts|)\]
where \(k\) specifies the number of Broad-View Frontiers (i.e., the number of coarse directions we want). We then apply K-means on the 2D coordinates of the frontier points, dividing them into \(k\) spatial clusters. Very small clusters (size $<\tau_{\text{size}}^{(0)}$) are removed to avoid noisy or unstable directions.

For each remaining cluster $\mathcal{C}_b$, we convert its points into polar angles around the robot and compute one representative direction by averaging these angles on the unit circle. If all clusters are removed, we fallback to uniformly partitioning the angular domain into \(k\) equal bins. The resulting set $\bvf=\{\theta_b\}$ forms the Broad-View Frontiers (BVFs), providing a few stable, high-level exploration directions.    

\vspace{0.2cm} \noindent \textbf{Close-up-View Frontier (CVF) Construction.}
BVFs provide coarse directions, but each direction may still span a wide region. To obtain more precise choices, we construct Close-up-View Frontiers (CVFs) inside each BVF. If a BVF cluster $\mathcal{C}_b$ contains fewer than three points, it is directly treated as a single CVF with direction $\theta_{b,0}=\theta_b$. Otherwise, we take the angular unit-vector embeddings of points in $\mathcal{C}_b$ and run a second K-means with
\[
k_b=\min(3,|\mathcal{C}_b|),
\]
where \(k_b\) is the number of finer CVFs to extract within this BVF. This reveals local angular structure in the neighborhood of that coarse direction. Sub-clusters smaller than $\tau_{\text{size}}^{(1)}$ are removed.

For each remaining sub-cluster $\mathcal{S}_{b,j}$, its direction $\theta_{b,j}$ is obtained by averaging its embedded unit vectors. The final CVF set is $\cvf=\{(\theta_{b,j},b)\}$, where each CVF keeps the index of its parent BVF. During action selection, the agent first chooses a BVF from $\bvf$ and then selects one of its CVFs in $\cvf$, enabling a simple and stable coarse-to-fine frontier selection process.   

\vspace{0.2cm} \noindent \textbf{Hierarchical Frontier Selection.}
For every direction $\theta$, we select a specific frontier cell that best represents this direction. 
Concretely, we choose the frontier cell whose bearing from the robot position $c$ is closest to $\theta$:
\[
\mathbf{p}^\ast(\theta)
=
\arg\min_{\mathbf{p}_i\in \fronts}
\bigl|\operatorname{wrap}(\operatorname{atan2}(p_i^y-c^y,\ p_i^x-c^x)-\theta)\bigr|.
\]
The simulator then renders an RGB view from $c$ while orienting the camera toward $\mathbf{p}^\ast(\theta)$.
These rendered images constitute the frontier snapshots used by the policy at both the BVF and CVF layers.
Basically, each abstract direction is turned into a real observation by pointing the camera toward the frontier cell that best matches that direction.

\begin{figure*}
    \centering
    \includegraphics[width=1\linewidth]{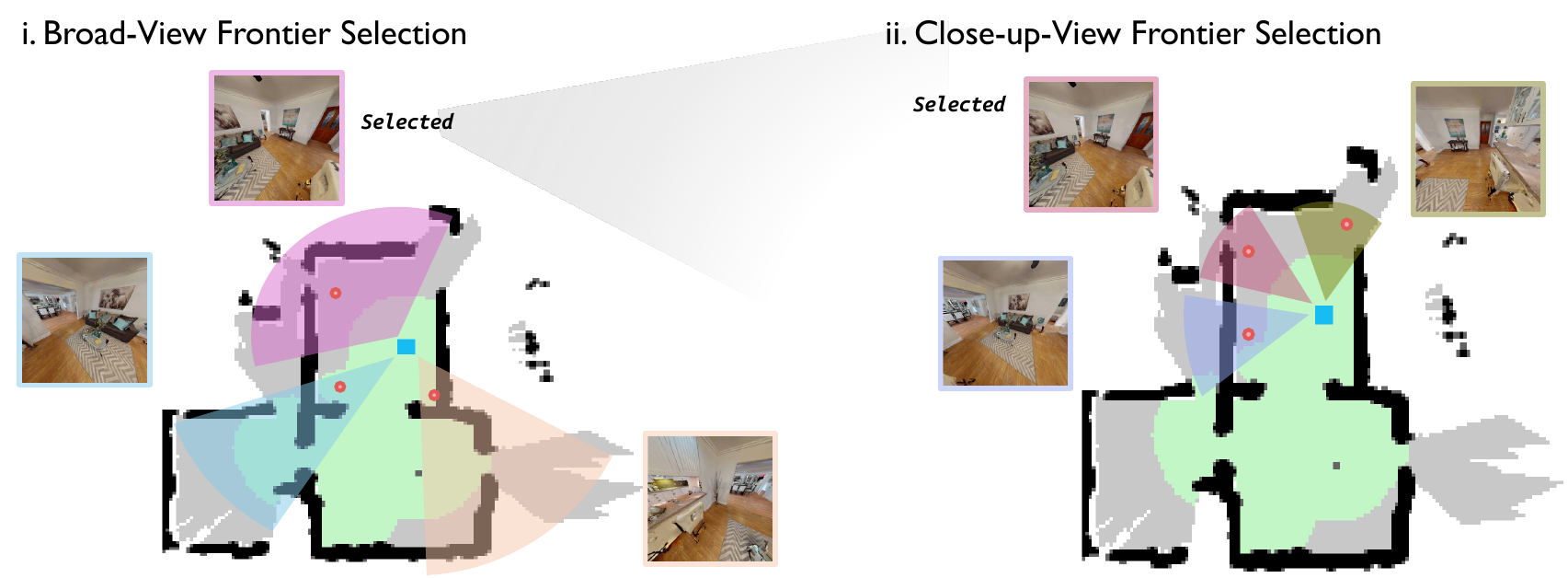}
    \caption{Hierarchical frontier selection using visualizations taken directly from \textbf{real exploration} data: the agent first picks a Broad-View frontier cluster (left), then selects a fine-grained frontier(close-up view) within that cluster (right)}
    \label{fig:hiera_example}
\end{figure*}

At decision time, the agent chooses an action through a coarse-to-fine procedure that mirrors the BVF $\rightarrow$ CVF hierarchy, and we also visualized it with a real example as shown in \cref{fig:hiera_example}.  
(1) The agent first considers only the Broad-View Frontiers (BVFs). For each BVF, we render an egocentric image looking toward its representative center. These rendered views, together with the question, local context, and optional retrospective experience, are assembled into a compact prompt. The LLM evaluates this prompt and selects the BVF whose direction appears most relevant for making progress on the task.
(2) After a BVF is chosen, the agent restricts attention to the Close-up-View Frontiers (CVFs) that belong to this parent BVF. A second prompt is constructed in the same style using only the CVF-level views, providing the LLM with fine-grained directional options within the selected region. The LLM outputs a final CVF choice, which we map back to the corresponding frontier cell in global coordinates. If a BVF contains no valid CVFs, or if the LLM returns an invalid index, we default to using the BVF’s representative point itself.
This hierarchical decomposition turns a large, unstable frontier set into a small, semantically structured action space. By first selecting a direction and then refining to a viewpoint, the model makes more reliable and interpretable decisions while preserving coverage of the underlying frontier geometry.
The detailed prompt is shown in \cref{sup:frontier_prompt}.

\subsection{Implementation of Embodied Exploration}\label{sup:explore}

This section describes the full implementation details of our embodied exploration pipeline, including the construction of working memory, the retrieval of retrospective experience, and their integration into the frontier-selection process. 

\vspace{0.2cm} \noindent \textbf{Implementation of Working Memory.}
Working memory provides a short-horizon textual summary of the agent’s recent progress within current exploration. At each step $t$, we gather the most recently \emph{chosen} frontier snapshots (typically the last five) and compress them into a short paragraph with a vision-language model. This paragraph describes (i) which regions were just explored, (ii) which cues were observed, and (iii) which directions remain uncertain. Crucially, the working memory is \emph{episode-bounded}: once the agent moves forward, older information is discarded and only the latest observations remain in the summary. During frontier ranking, the working memory is passed to the model alongside the question and candidate frontiers, ensuring that the agent does not treat each snapshot as an isolated frame but as part of a coherent exploration history.

\vspace{0.2cm} \noindent \textbf{Retrospective Experience Set Format.}
To supply long-horizon structure, we build an experience set from training episodes. Each training question corresponds to a completed exploration trajectory. For every trajectory, we store three components: (i) the question text; (ii) a set of stored frontier snapshots, each recorded at the time of frontier selection and associated with a trajectory index and step; and (iii) a two-paragraph trajectory abstraction summarizing typical patterns, useful strategies, and common failure modes. This abstraction is produced as described in the main text and serves as the reusable unit of experience during new episodess~\cite{tulving1985memory,sutton1998reinforcement}.

\vspace{0.2cm} \noindent \textbf{Details of Salient Experience Recall.}
During a new episode, retrospective experience must be tailored to the current state and the current frontier candidates. At step $t$, each candidate frontier in $\fronts$ is represented by a local RGB snapshot. We embed the snapshot using OpenCLIP ViT-H/14~\cite{Cherti_2023} and retrieve stored snapshots in the experience set with highest cosine similarity. Because each stored snapshot is linked to a trajectory  $\tau_{i}$, this step yields a ranked list of visually relevant trajectories.

In parallel, we compute task similarity. The current question $q_i$ is encoded with a lightweight sentence transformer (MiniLM-L6-v2)\cite{reimers2019sentencebertsentenceembeddingsusing}, and cosine similarity is evaluated against all stored questions in the set. Since every question corresponds to a trajectory, this produces a second ranked list of semantically relevant trajectories. Scene similarity captures “what the agent currently sees,” while task similarity captures “what the agent is currently asked to find.” Both routes are complementary and jointly determine which past experiences are informative.

To merge the two rankings, we apply Reciprocal Rank Fusion (RRF):
\[
\text{score}_{i}
= \frac{1}{k + \text{rank}^{(\mathrm{scene})}_{i}}
+ \frac{1}{k + \text{rank}^{(\mathrm{task})}_{i}},
\]
with constant $k=60$. This late-fusion rule is robust to scale differences across modalities and prioritizes trajectories that are jointly similar in appearance and meaning. We select the top-$K$ trajectories and gather their trajectory abstractions, concatenating them into a single replay context
\[
\mathcal{R}_t = \{\text{Abstraction}(\tau_i)\}_{i \in \text{Top-}K}.
\]
These abstractions do not instruct a specific action; instead, they offer distilled knowledge about how similar spatial layouts tend to unfold, where informative views usually appear, and what kinds of decisions often lead to redundant loops.

\vspace{0.2cm} \noindent \textbf{Hyperparameter.} In the hierarchical frontier selection module, both in Broad- and Close-up-View will produce three candidate directions through a fixed three-way k-means clustering step. This yields up to three BVFs and up to three CVFs per BVF, giving the agent a stable and compact set of choices at every decision point.

For salient experience recall, we recall the top five retrospective experiences from the set using the fused scene and task similarity scores. These settings remain unchanged across all experiments in \explorer.

\vspace{0.2cm} \noindent \textbf{Contextual Replay with Frontier Selection.}
At every step of exploration, the agent selects a frontier using a unified prompt that contains all available textual and visual information. The model is not given any special interface for working memory or retrospective experience; instead, both forms of context are injected as independent textual blocks and interpreted directly within the same reasoning process.

\emph{Working memory} provides a short-horizon summary of the agent’s recent behavior. It lists which areas were inspected, which views yielded useful or redundant information, and which directions remain unexplored. This allows the model to maintain local continuity and avoid revisiting regions that have already been covered.

\emph{Retrospective experience} contributes long-horizon structure. The retrieved trajectory abstractions summarize how similar situations unfolded in past episodes, including typical region transitions, early signs of dead ends, functional cues, and high-yield search patterns. These abstractions do not prescribe a specific action but act as strategic priors that shape how the current observations are interpreted.

During inference, the agent receives:  
(1) the question,  
(2) the candidate frontiers (images),  
(3) the working-memory summary, and  
(4) the replay abstractions relevant to the current step.  
All four components are placed into the same decision prompt, where the model is instructed to read them jointly and justify its choice.

The key idea is that the model is free to combine these signals through natural-language reasoning. Working memory anchors the decision to the current episode’s partial exploration status, ensuring short-term consistency, while retrospective experience introduces transferable knowledge that biases the model toward historically informative directions. This fusion requires no modification to the navigation system or mapping pipeline; it is achieved entirely through prompt design and unified textual conditioning.

\subsection{Technical Details of Benchmarks}
\label{sec:benchmarks}


\vspace{0.2cm} \noindent \textbf{Question Subset and Data Split.} 
We follow the 3D-Mem configuration and evaluate only on the 41 \emph{A-EQA} questions provided by that work, which cover five HM3D scenes. The remaining 164 questions come from eleven different scenes and are used solely to produce retrospective experience: we run complete episodes on these scenes and convert each trajectory into an abstraction used later as a textual prior. These abstractions do not contribute actions, states, or working memory during evaluation. All test episodes use the official A-EQA scene assignments, starting poses, and scoring scripts, and each question is executed once in a freshly reset environment. Scenes that cannot load a valid navigation mesh or are marked invalid in A-EQA are removed without further modification. During evaluation, the model receives the question, the online working-memory summary, the current frontier candidates, and the retrieved retrospective abstractions, with all methods sharing the same backbone and interaction settings.

\emph{GOAT-Bench} is constructed from HM3DSem \cite{yadav2023habitat_hm3dsem} and comprises 136 scenes in the training split and 36 scenes in the val-unseen split. Each scene contains 10 episodes, and every episode includes 5-10 target objects specified by either an object category, a language description, or an image.
Due to the large scale of the val-unseen split, we follow 3D-Mem and use only the first episode of each of the 36 scenes as our evaluation set, resulting in 278 subtasks, treating each queried object as an individual subtask, analogous to one question in AEQA.
The training split is used to construct retrospective experience. Specifically, we roll out the experience-free frontier exploration policy on the first episode of 34 training scenes, covering 255 subtasks. For each rollout, we prompt the MLLM to summarize the trajectory, reflect on how each action contributed to the final outcome, and abstract the key lessons learned. These distilled summaries constitute our retrospective experience set.
During evaluation, the model is queried with modality-dependent questions of the form: ``Can you find the {category}?'', ``Can you find the object described as {language description}?'', or ``Can you find the object captured in the following image? \{image\}''. As in A-EQA, exploration is enhanced with online working memory, frontier candidates, and retrieved salient retrospective abstractions. We additionally adapt the prompt used in A-EQA to support the navigation setting. Unlike A-EQA, where the environment resets after each question, the agent continues exploration across subtasks within the same episode, sequentially locating each target until all objects are found.

\section{Extended Experimental Study}
\label{sup:extended}
\subsection{Extended Results on A-EQA}

\begin{figure}
    \centering
    \includegraphics[width=1\linewidth]{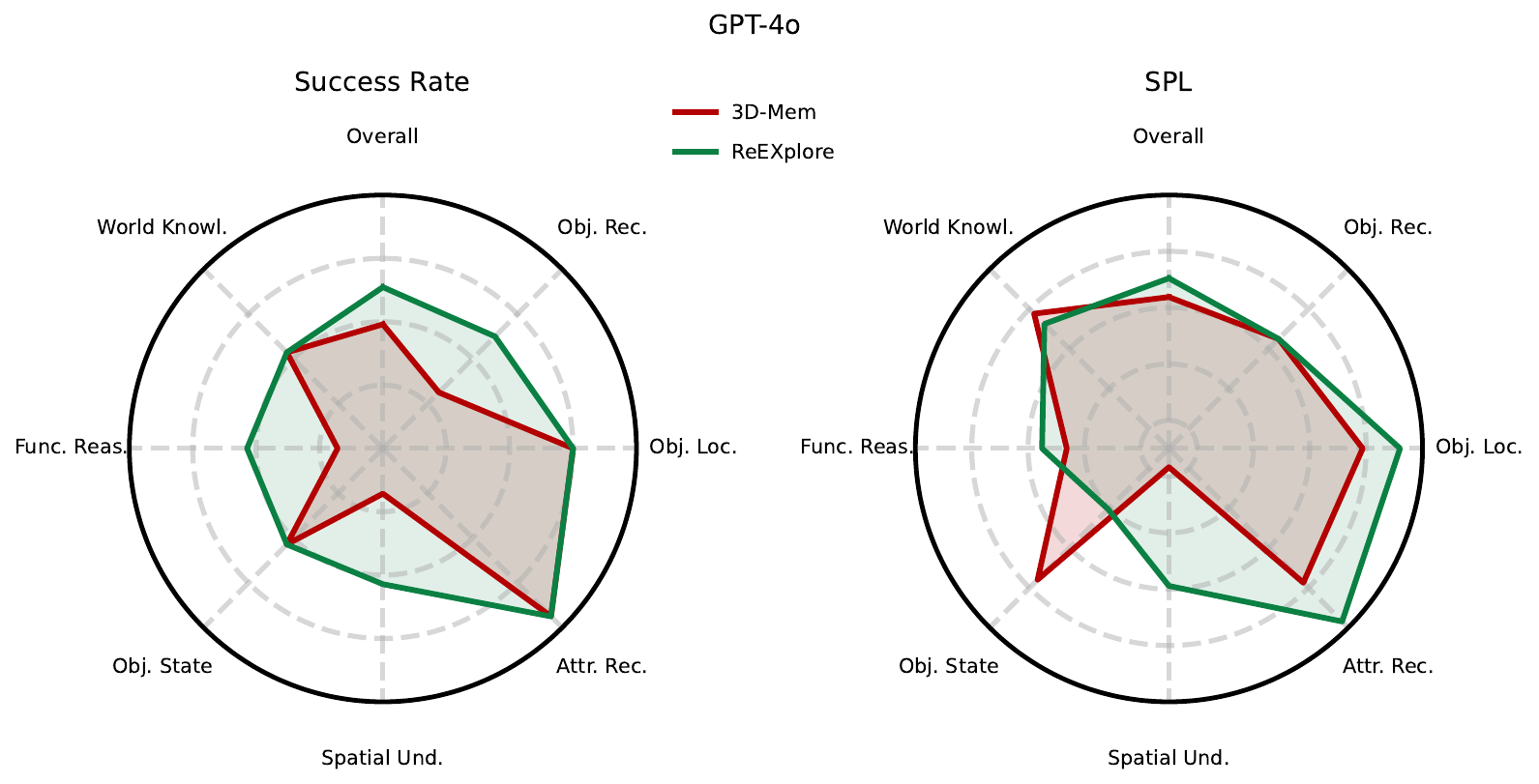}
    \caption{Performance of \explorer (with \gpt) on A-EQA by question categories, showing both Success Rate (Succ.) and SPL. Our approach outperms the strongest base-line model by a large margin in all task categories.}
    \label{fig:radargpt}
\end{figure}

\cref{tab:succ_spl} is the full performance of \explorer (with \qwen and \gpt) on A-EQA by question categories, showing both Success Rate (Succ.) and SPL. Our approach outperms the strongest baseline model by a large margin in all task categories. The result of \gpt also show this trend in \cref{fig:radargpt}.

\begin{table*}[htbp]
\centering
\scriptsize
\adjustbox{max width=\textwidth}{
\begin{tabular}{lcccccccccccccccc}
\toprule
\textbf{Method} &
\multicolumn{2}{c}{\textbf{object recog.}} &
\multicolumn{2}{c}{\textbf{object loc.}} &
\multicolumn{2}{c}{\textbf{attribute recog.}} &
\multicolumn{2}{c}{\textbf{spatial und.}} &
\multicolumn{2}{c}{\textbf{object state recog.}} &
\multicolumn{2}{c}{\textbf{functional reas.}} &
\multicolumn{2}{c}{\textbf{world knowl.}} &
\multicolumn{2}{c}{\textbf{overall}} \\
\cmidrule(lr){2-3} \cmidrule(lr){4-5} \cmidrule(lr){6-7} \cmidrule(lr){8-9}
\cmidrule(lr){10-11} \cmidrule(lr){12-13} \cmidrule(lr){14-15} \cmidrule(lr){16-17}
& \textbf{Succ.} & \textbf{SPL} & \textbf{Succ.} & \textbf{SPL} & \textbf{Succ.} & \textbf{SPL} &
\textbf{Succ.} & \textbf{SPL} & \textbf{Succ.} & \textbf{SPL} & \textbf{Succ.} & \textbf{SPL} &
\textbf{Succ.} & \textbf{SPL} & \textbf{Succ.} & \textbf{SPL} \\
\midrule
\multicolumn{17}{l}{\textit{Commercial Models}} \\
\textbf{3D-Mem (\gpt)} & 62.5 & 62.5 & 80.0 & 69.4 & 87.5 & 68.7 & 57.1 & 38.3 & 71.4 & 68.0 & 57.1 & 53.1 & 71.4 & 68.8 & 69.6 & 61.9 \\
\explorer (\gpt) & 75.0 & 62.5 & 80.0 & 76.0 & 87.5 & 78.5 & 71.4 & 59.4 & 71.4 & 50.3 & 71.4 & 57.5 & 71.4 & 66.3 & 75.5 & 65.2 \\
\midrule
\multicolumn{17}{l}{\textit{Open-source Models}} \\
\textbf{3D-Mem (\qwen)} & 62.5 & 49.4 & 60.0 & 37.0 & 50.0 & 38.8 & 14.3 & 14.3 & 42.9 & 30.5 & 57.1 & 38.5 & 57.1 & 43.8 & 50.9 & 36.4 \\
\explorer (\qwen) & 75.0 & 60.6 & 70.0 & 58.0 & 62.5 & 60.5 & 42.9 & 41.8 & 42.9 & 32.5 & 57.1 & 57.1 & 71.4 & 55.3 & 58.5 & 52.9 \\
\bottomrule
\end{tabular}
}
\caption{\textbf{Performance on A-EQA\cite{majumdar2024openeqa} by question categories}, showing both Success Rate (Succ.) and SPL. }
\label{tab:succ_spl}
\end{table*}

\subsection{Extended Results on GOAT-Bench}
\label{sup:goatbench_result}

\cref{tab:goatbench-full} presents the detailed performance of \explorer on GOAT-Bench across different query modalities, including object categories, language descriptions, and images. When using an open-source MLLM such as \qwen, our method achieves higher success rates than the 3D-Mem baseline in both the object-category and image modalities, with a particularly notable 25\% improvement for image queries, resulting in an overall gain of nearly 4\% in success rate. The SPL improvements are even more pronounced, showing consistent boosts across all modality types. For object-category queries, the SPL increases by as much as 82\%. However, when deployed with \gpt, our approach falls slightly behind the strongest baseline but still performs comparably to the second-best methods. We attribute this gap primarily to the limited budget and resources available for prompt tuning.

\begin{table*}[t]
\centering
\scriptsize
\setlength{\tabcolsep}{3.4mm}
\begin{tabular}{l cc cc cc cc}
\toprule
& \multicolumn{2}{c}{\textbf{Object Category}} &
  \multicolumn{2}{c}{\textbf{Language}} &
  \multicolumn{2}{c}{\textbf{Image}} &
  \multicolumn{2}{c}{\textbf{Overall}} \\
\cmidrule(lr){2-3}\cmidrule(lr){4-5}\cmidrule(lr){6-7}\cmidrule(lr){8-9}
\textbf{Method} & \textbf{Success Rate} & \textbf{SPL} &
\textbf{Success Rate} & \textbf{SPL} &
\textbf{Success Rate} & \textbf{SPL} &
\textbf{Success Rate} & \textbf{SPL} \\
\midrule
\multicolumn{9}{l}{\textit{\textbf{Open-Sourced MLLM Exploration}}}\\
3D-Mem (\qwen) & 60.3 & 21.6 & \textbf{47.6} & 19.5 & 39.0 & 21.0 & 49.4 & 20.7 \\   
\explorer (\qwen) & \textbf{64.6} & \textbf{39.3} & 45.1 & \textbf{28.2} & \textbf{48.9} & \textbf{29.5} & \textbf{53.2} & \textbf{32.6} \\    
\midrule
\multicolumn{9}{l}{\textit{\textbf{\gpt Exploration}}} \\
Explore-EQA*                & 64.7 & \underline{48.4} & 42.9 & 22.7 & 56.8 & 41.8 & 55.0 & 37.9 \\
CG w/ Frontier Snapshots*   & 65.3 & 44.7 & \underline{55.0} & 38.9 & \underline{64.0} & \underline{52.8} & \underline{61.5} & \underline{45.3} \\
3D-Mem w/o memory*          & \textbf{69.9} & 45.4 & 50.3 & 30.1 & 54.4 & 39.5 & 58.6 & 38.5 \\
3D-Mem & \underline{66.7} & \textbf{49.1} & \textbf{59.3} & \textbf{45.0} & \textbf{69.3} & \textbf{54.0} & \textbf{65.1} & \textbf{49.3} \\    
\explorer & 65.7 & 46.2 & 54.9 & \underline{39.2} & 58.0 & 44.9 & 59.8 & 42.5 \\    
\bottomrule
\end{tabular}
\caption{\textbf{Performance on GOAT-Bench\cite{yang20253d} by question modalities}, evaluated on subset of the “Val Unseen” split. “CG” denotes ConceptGraphs\cite{gu2024conceptgraphs}. \textbf{Bold} indicates the best results, \underline{underlined} values denote the second-best. Methods with * are reported from and 3D-Mem\cite{yang20253dmem3dscenememory}. }
\label{tab:goatbench-full}
\end{table*}

\subsection{Significance of Retrospective Experience}
\begin{figure}
    \centering
    \includegraphics[width=1\linewidth]{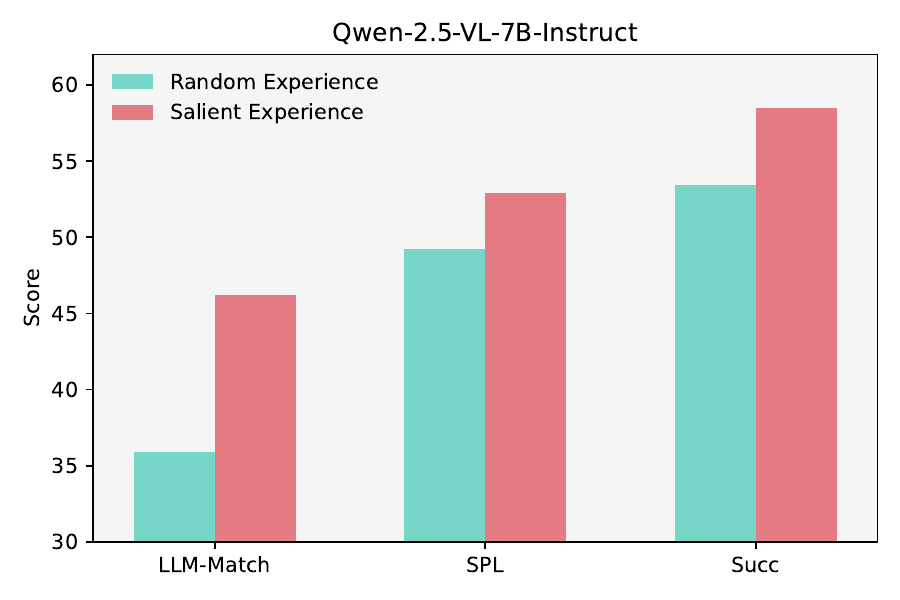}
    \caption{Performance when using between top-five salient recall and random pick. Both using 5 retrospective experiences.}
    \label{fig:random_sim}
\end{figure}

\vspace{0.2cm} \noindent \textbf{Comparing Different Retrieval  Methods.}
To evaluate the reliability of our salient experience recall mechanism, we compare the proposed salient recall method against a random retrieval (shown in \cref{fig:random_sim}). In this setting, both approaches supply the agent with 5 pieces of experiences at each step, differing only in how the trajectories are selected. Across all evaluation metrics, salient recall consistently improves overall performance, indicating that retrieving experience based on scene similarity and task relevance provides more useful guidance than sampling trajectories at random.

\vspace{0.2cm} \noindent \textbf{Quality Assessment of Retrospective Experience.}
To examine how the quality of retrospective experience affects downstream performance, we evaluate every retrospective abstraction using \qwen as an LLM-judge. Each abstraction is scored along four complementary dimensions defined as:

\begin{itemize}
    \item \textbf{Generality}: whether the abstraction captures transferable exploration principles rather than recounting a single episode.
    \item \textbf{Relevance}: whether it aligns with the original question and highlights cues directly related to answering it.
    \item \textbf{Conciseness}: whether the description conveys key ideas clearly without redundancy or digressions.
    \item \textbf{Actionability}: whether it provides concrete, environment-grounded hints that could guide future exploration decisions.
\end{itemize}

\begin{figure}[t]
    \centering
    \begin{subfigure}{1.0\linewidth}
        \centering
        \includegraphics[width=0.86\linewidth]{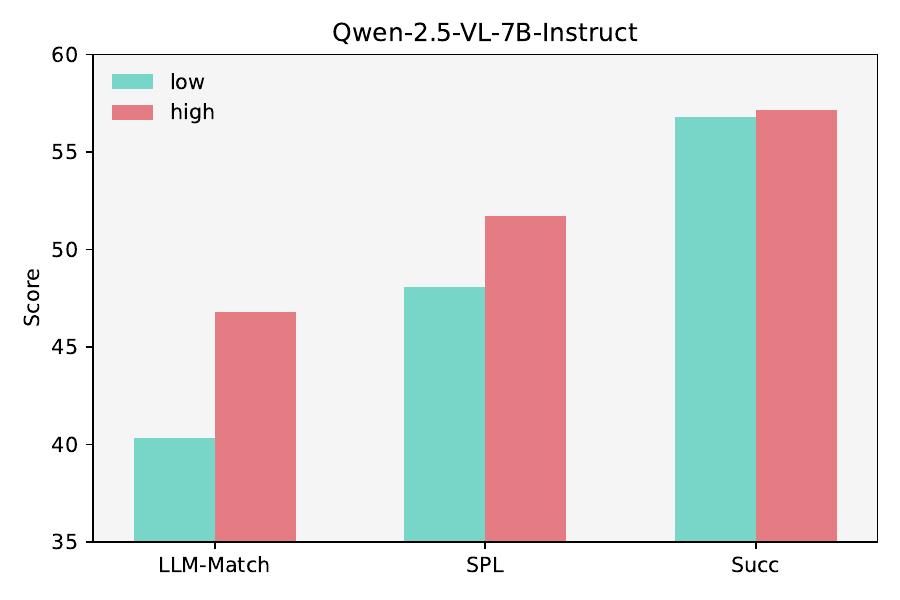}
        \caption{}
        \label{fig:llmscore_qwen}
    \end{subfigure}

    \vspace{0.2cm}

    \begin{subfigure}{0.86\linewidth}
        \centering
        \includegraphics[width=\linewidth]{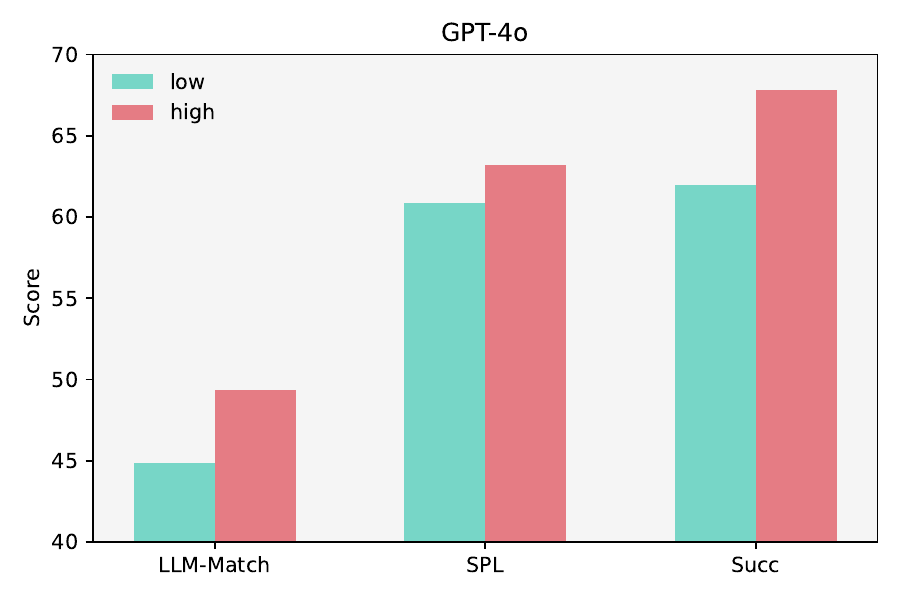}
        \caption{}
        \label{fig:llmscore_gpt}
    \end{subfigure}

    \caption{\explorer performance when using different quality levels of retrospective experiences (high quality vs.\ low quality) judged by MLLM itself. We notice that high quality experience does improve the model performance.}
    \label{fig:llmscore_combined}
\end{figure}

\noindent
For each dimension, \qwen produces a 1--5 score together with a brief justification; the four scores are averaged to obtain an overall quality score for that abstraction. Based on this score, we divide all abstractions into \emph{high-quality} and \emph{low-quality} groups, and perform replay-based inference using only the abstractions from one group at a time. As shown in \cref{fig:llmscore_qwen,fig:llmscore_gpt}, high-quality abstractions consistently provide stronger guidance and lead to substantially better exploration performance, demonstrating that the effectiveness of retrospective experience depends strongly on its intrinsic quality.

\begin{figure}
    \centering
    \includegraphics[width=0.86\linewidth]{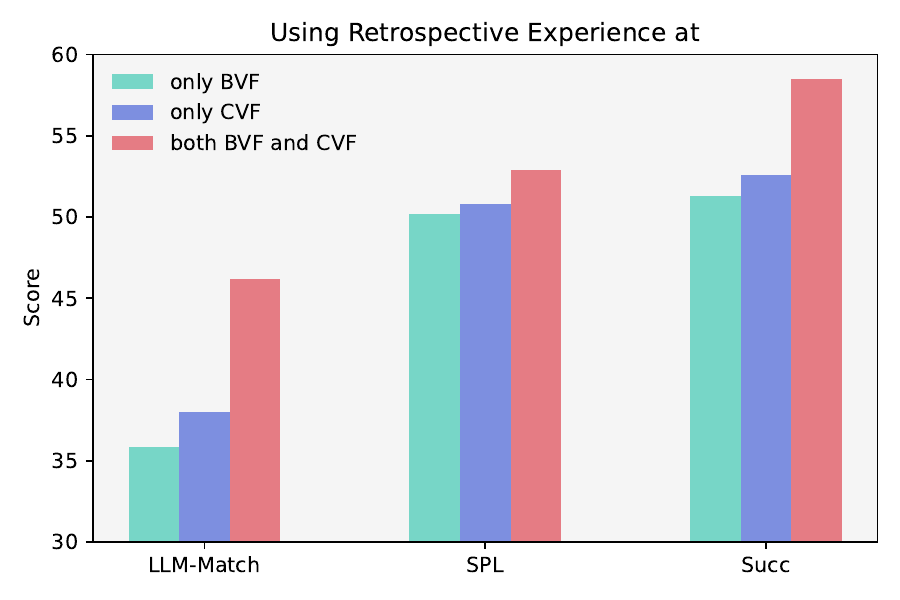}
    \caption{\explorer (w/ \qwen) performance when retrospective experience is replayed at different stages of hierarchical frontier selection: only at the Broad-View Frontier (BVF) selection, only at the Close-up-View Frontier (CVF) Selection, or at both stages (as in \explorer).}
    \label{fig:bvfcvf}
\end{figure}

\vspace{0.2cm} \noindent \textbf{Effects of Experience Replay in the Hierarchical Frontier Selection Stages.}
Our hierarchical frontier selection contains two stages: a coarse \emph{Broad-View} stage and a finer \emph{Close-up-View} stage. To examine how retrospective experience interacts with this hierarchy, we perform controlled experiments where trajectory abstractions are injected into only one layer at a time. As shown in \cref{fig:bvfcvf}, using retrospective experience at the Broad-View Selection alone produces a different behavior pattern from using it only at the Close-up-View Selection. When abstraction are provided to \emph{both} layers, however, the agent achieves the strongest overall performance. This comparison suggests that retrospective experience influences exploration decisions at multiple levels of granularity, and its full benefit emerges when both directional selection and local frontier evaluation are informed by the retrieved abstractions.

\begin{figure}
    \centering
    \includegraphics[width=1\linewidth]{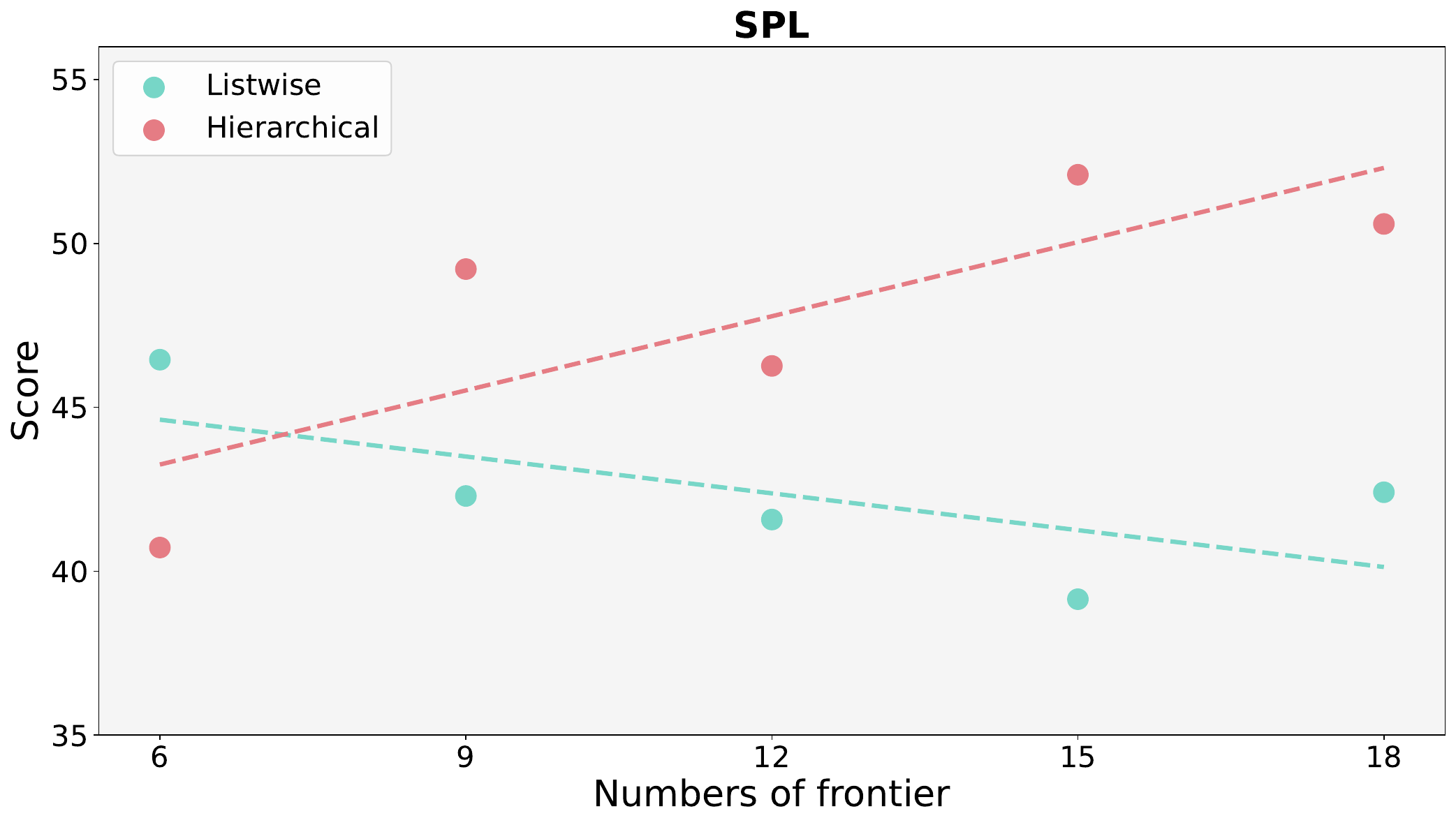}
    \caption{SPL Performance comparison between listwise and hierarchical frontier selection as the number of frontier candidates increases(using \qwen).}
    \label{fig:listhiera}
\end{figure}

\begin{figure*}[htbp]
    \centering

    \begin{subfigure}{0.95\linewidth}
        \centering
        \includegraphics[width=\linewidth]{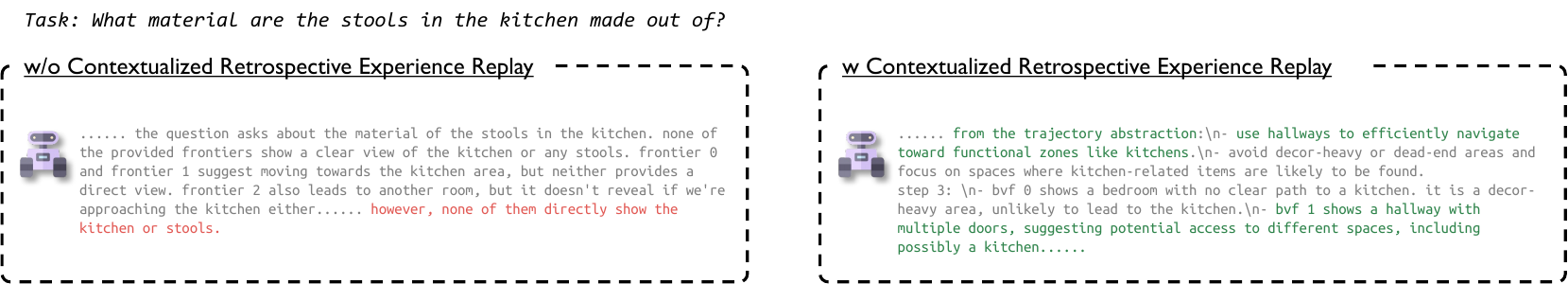}
        \caption{Example 1.}
        \label{fig:case811}
    \end{subfigure}

    \vspace{0.5em}

    \begin{subfigure}{0.95\linewidth}
        \centering
        \includegraphics[width=\linewidth]{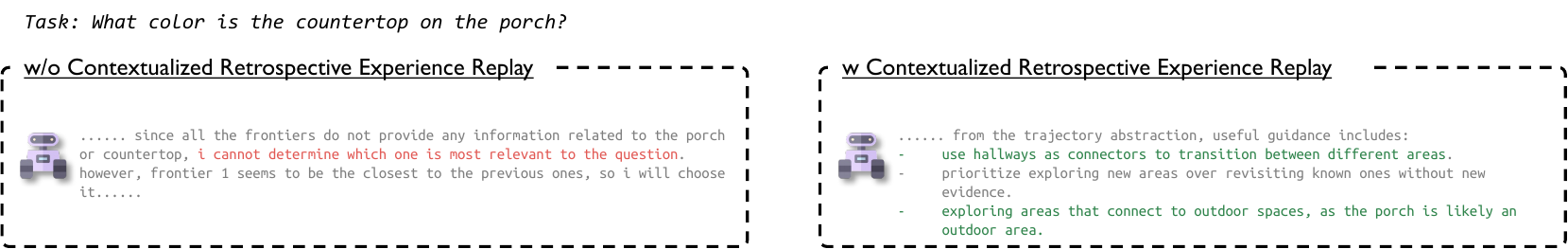}
        \caption{Example 2.}
        \label{fig:case812}
    \end{subfigure}

    \caption{Qualitative examples of how contextualized retrospective experience replay impact the reasoning traces of frontier selection. We highlight failed reasoning trace in \textcolor{BrickRed}{red} and successful ones in \textcolor{ForestGreen}{green}.}
    \label{fig:case81}
\end{figure*}

\subsection{Comparing Different Frontier Selection Methods on Open-Sourced Models}

To study how various frontier selection formulations influence exploration quality, we evaluate four alternative methods: listwise, pointwise, pairwise, and hierarchical, under the \qwen backbone. All methods share the same observations, question prompts, and action budgets; they differ only in how candidate frontier snapshots are organized and presented to the VLM for decision making. The results are shown in \cref{tab:diffwise}.

\vspace{0.2cm} \noindent \textit{Listwise Frontier Selection.}
Listwise frontier selection provides a natural reference point: all frontier snapshots are presented simultaneously in a single prompt, and the VLM is asked to choose one based on the question. This enables global comparison across all candidates, allowing the model to reason about relative informativeness and redundancy. However, when many frontiers appear at once, the prompt can become visually dense, making it harder for the model to maintain focus or utilize its context budget effectively.

\vspace{0.2cm}
\noindent
\textit{Pointwise Frontier Selection.}
Pointwise frontier selection considers each frontier independently. The VLM receives only one candidate at a time and decides whether exploring it could help answer the question. This produces clear and interpretable decisions but does not account for relationships among candidates, which may lead to inconsistent preferences when several frontiers share similar cues.

\vspace{0.2cm}
\noindent
\textit{Pairwise Frontier Selection.}
Pairwise frontier selection introduces direct comparison between two frontiers at a time. The VLM must choose one and justify its choice, and this procedure is repeated in a tournament-style fashion until a final option remains. This approach supports more stable relative preferences than pointwise evaluation, but it is computationally more expensive and can be sensitive to the order of comparisons.

\vspace{0.2cm}
\noindent
\textit{Hierarchical Selection.}
Our hierarchical approach organizes frontiers into two layers: a coarse directional layer and a finer local layer. The VLM first chooses a coarse direction and then selects a specific frontier within that region. This coarse-to-fine decomposition reduces visual clutter, preserves spatial coherence, and aligns with how indoor environments naturally structure navigational choices. It allows the model to reason at the appropriate granularity rather than treating all frontiers as equally independent options.

As shown in \cref{tab:diffwise}, the hierarchical frontier selection achieves the best performance across all metrics. While listwise, pointwise, and pairwise methods offer different ways to handle frontier candidates, none captures spatial structure as effectively as the hierarchical formulation. By separating global directional reasoning from local view selection, the hierarchical method enables more stable, coherent, and ultimately more successful exploration behavior on open-sourced VLMs.

\vspace{0.2cm} \noindent \textbf{Impact of More Frontier Candidates.}
To better highlight the performance difference between listwise and hierarchical frontier selection, we vary the number of frontier snapshots presented to the model at each step. This is achieved by adjusting the clustering granularity, which directly controls the number of candidates the VLM must evaluate. As shown in \cref{fig:listhiera}, when the candidate set is small (e.g., six frontiers), listwise selection holds a slight advantage, as the global comparison remains easy for the model. However, as the number of candidates increases, the performance gap reverses: hierarchical selection quickly surpasses listwise and continues to widen, with the largest margin observed when fifteen frontiers are presented. This pattern indicates that hierarchical selection is more robust when the model is required to handle richer and more complex visual sets, while listwise selection deteriorates under heavier input load. The comparison demonstrates that hierarchical structuring of frontier selection provides more reliable and efficient guidance for exploration, especially as perceptual complexity grows.

\begin{table}[t]
\centering
\scriptsize
\adjustbox{max width=\columnwidth}{
\begin{tabular}{ccccc}
\toprule
\multicolumn{1}{c}{\textbf{Method}} 
& \multicolumn{4}{c}{\textbf{Overall}} \\
& \textbf{\llm} 
& \textbf{\llmspl} 
& \textbf{SPL} 
& \textbf{Succ.} \\
\midrule

listwise      & 34.2 & 15.9 & 45.1 & 56.2 \\
pointwise     & 38.5 & 23.3 & 44.6 & 52.6 \\
pairwise      & 39.4 & 23.1 & 46.9 & 54.1 \\
hierarchical  & 42.1 & 26.7 & 48.2 & 57.7 \\

\bottomrule
\end{tabular}
}
\caption{Results of different frontier selection strategies with  \qwen backbone.}
\label{tab:diffwise}
\end{table}

\section{Qualitative Analysis}
\label{sup:qual_analysis}
\subsection{Impact of Retrospective Experience}
\Cref{fig:case81} illustrates how retrospective experience replay alters the agent’s reasoning during frontier selection. In both examples, the baseline MLLM agent (\emph{w/o replay}) struggles because none of the visible frontiers directly reveal the queried object or region. As a result, its decisions become either uncertain or arbitrary. For example, selecting frontiers simply because they appear ``closest'' or because none seem clearly relevant.

With contextualized replay (\emph{w/ replay}), the agent instead grounds its reasoning in distilled experience abstractions retrieved from past, similar episodes. These abstractions provide high-level strategic cues, such as navigating through hallways to access functional zones like kitchens, avoiding decor-heavy dead-ends, prioritizing unexplored connectors, or preferring regions that lead to potential outdoor areas. Guided by these priors, the agent generates more targeted and confident frontier choices, even when direct visual evidence is absent.

Overall, these examples show that retrospective experience replay does not merely add textual hints, but reshapes the agent’s decision trace: from reactive, view-based navigation to proactive, strategy-driven exploration aligned with functional layouts of indoor environments.

\subsection{Impact of Hierarchical Frontier Selection}
\begin{figure*}[!t]
    \centering

    \begin{subfigure}{0.92\linewidth}
        \centering
        \includegraphics[width=\linewidth]{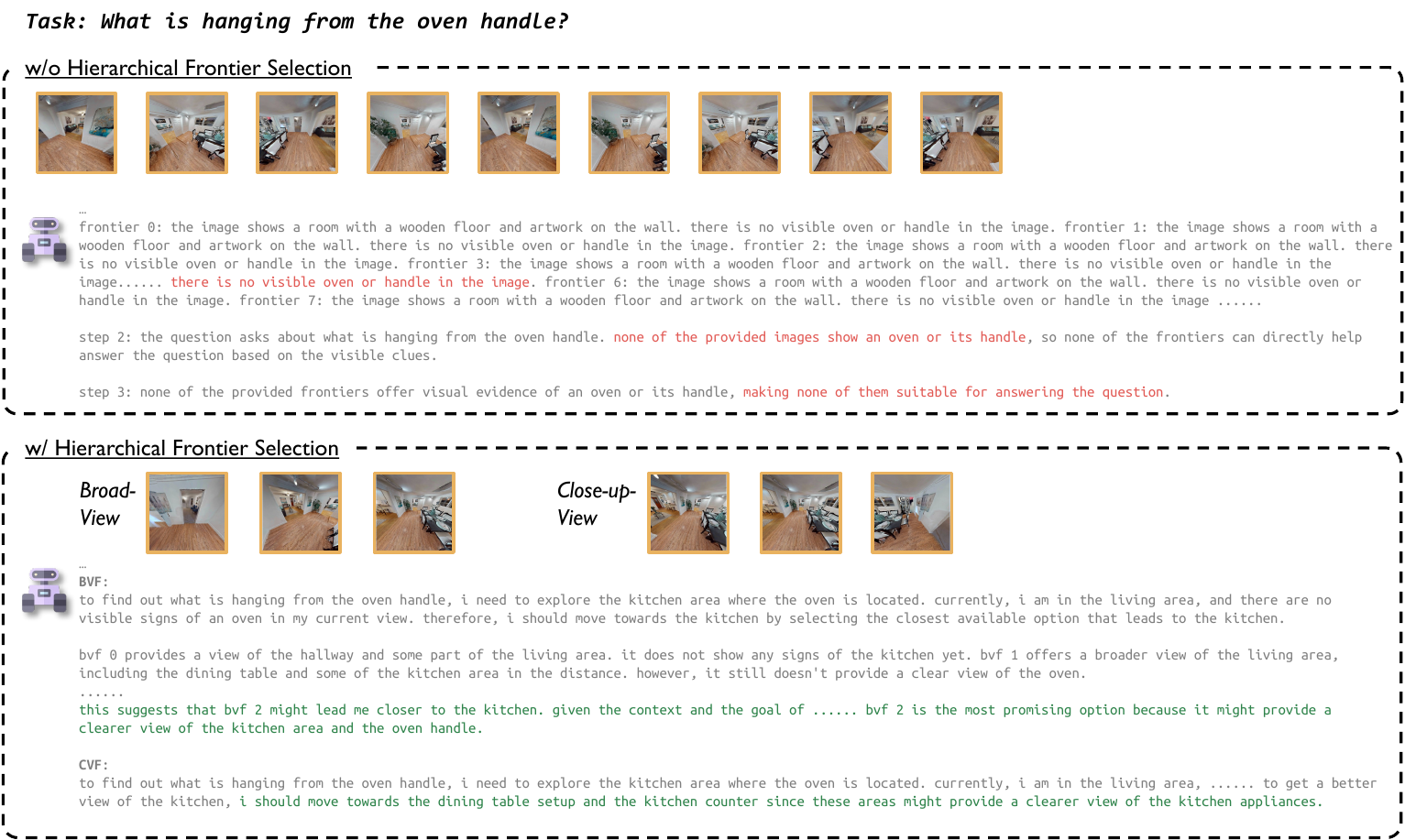}
        \caption{Example 1.}
        \label{fig:case821}
    \end{subfigure}

    \vspace{0.5em}

    \begin{subfigure}{0.92\linewidth}
        \centering
        \includegraphics[width=\linewidth]{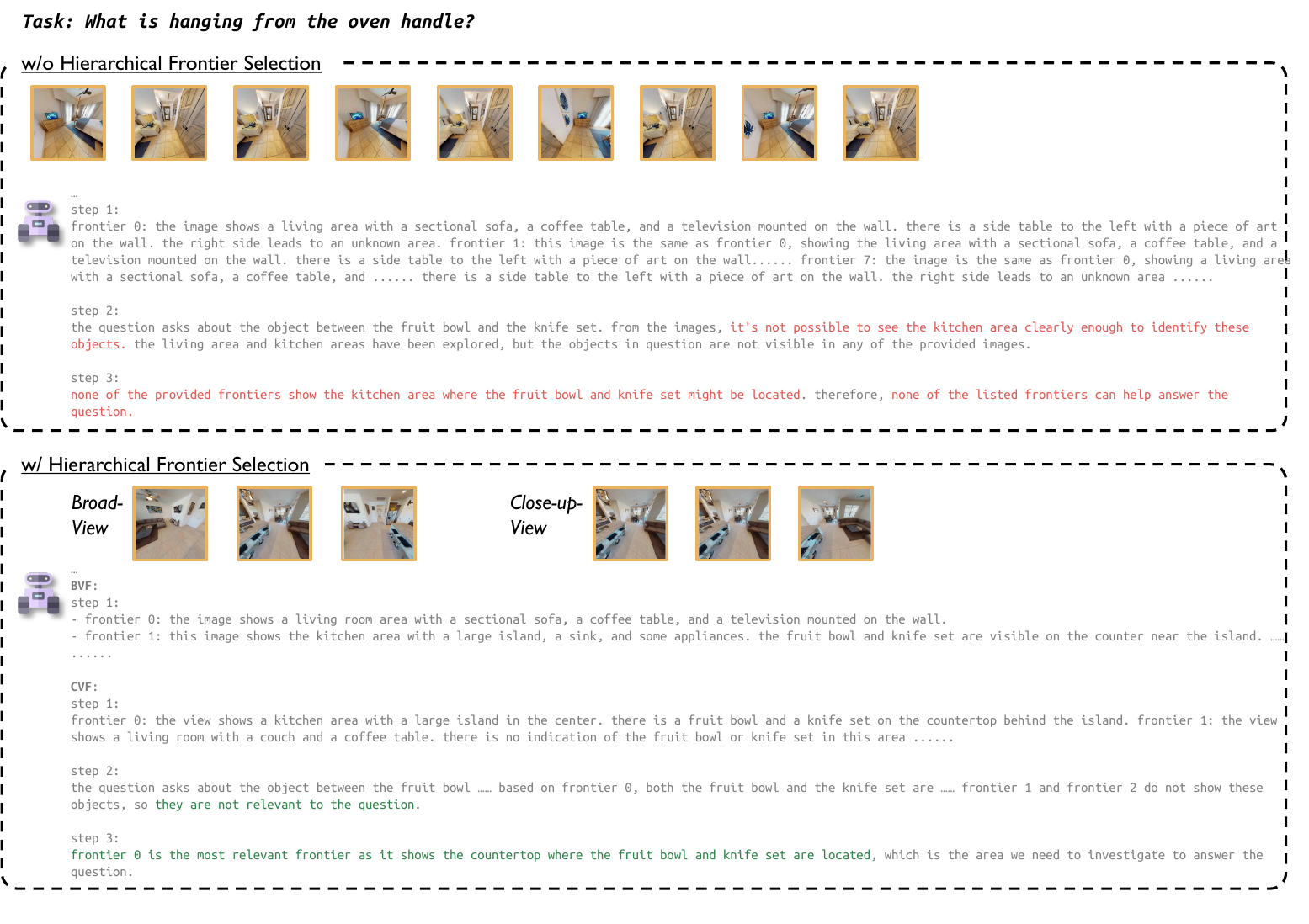}
        \caption{Example 2.}
        \label{fig:case822}
    \end{subfigure}

    \caption{Qualitative examples of how hierarchical frontier selection outperforms other frontier selection baseline (listwise ranking).We highlight failed reasoning trace in \textcolor{BrickRed}{red} and successful ones in \textcolor{ForestGreen}{green}.}
    \label{fig:case82}
\end{figure*}

\Cref{fig:case82} illustrates how hierarchical frontier selection enables more reliable and question-directed exploration compared to a flat frontier-ranking baseline. In both examples, the non-hierarchical agent struggles because all candidate frontiers appear visually similar and none directly reveal the target object, leading to indecisive reasoning and suboptimal choices.

With hierarchical frontier selection, the agent first performs broad-view reasoning to identify frontier regions that are more likely to contain functionally relevant spaces (e.g., kitchen areas when searching for oven handles, or regions connected to countertops when locating fruit bowls or knife sets). It then applies close-up reasoning to refine the choice within the selected region. This coarse-to-fine process yields a clearer interpretation of frontier cues and a more grounded selection that aligns with the underlying task semantics.

Across both examples, hierarchical frontier selection helps the agent filter out irrelevant regions, prioritize informative ones, and select frontiers that lead to faster exploration.

\begin{SelfReflectionBox*}[title={Retrospective Experience Abstraction}]
\footnotesize
You are a self-reflective embodied exploration agent. Your goal is to produce a two-part analysis of a complete exploration trajectory through \textbf{REFLECTION} and \textbf{ABSTRACTION}.

\begin{itemize}
  \item \textbf{Input Schema:}
    \begin{itemize}
      \item \textbf{Target Task:} \{question\_text\}
      \item \textbf{Exploration Trajectory:} a caption-style textual description summarizing the agent's exploration path, visited regions/rooms, transitions, and key observations
      \item \textbf{Final Outcome:} \{task\_outcome\}
    \end{itemize}

  \item \textbf{Output Format (must be exact and ordered):}
    \begin{itemize}
      \item \textbf{REFLECTION:} your output \emph{must} contain exactly five labeled blocks in this order:
      \begin{itemize}
        \item \textbf{Step 0 (Task Understanding) --- 2--3 sentences:}  
        Paraphrase succinctly what the question asks (e.g., find/verify/compare), and what constitutes success.

        \item \textbf{Step 1 (Trajectory) --- 8--10 sentences:}  
        Summarize the overall trajectory across the captioned exploration.  
        Describe the entry points and the major regions/rooms traversed (e.g., entrance, hallway, kitchen zone, utility area, living space), and the key transitions between them.  
        Indicate movement directionality (toward/away from salient regions or landmarks) and explain why the route changed (e.g., encountering new evidence or exhausting an area).  
        Focus on path logic and coverage (what was visited first/next/last), not on per-image details.

        \item \textbf{Step 2 (Env--Object Associations) --- 4--6 sentences:}  
        Provide general priors linking categories to regions.  
        Use generic categories and regions (e.g., signage near entrances/hubs; cookware in kitchen-like areas; cleaning supplies near sinks/utility corners; clothing/linens near bedroom/closet zones).  
        Avoid scene-specific item names.

        \item \textbf{Step 3 (Strategy $\times$ Question Type + Directional Priors) --- 4--6 sentences:}  
        Give concrete guidance per question type with directional priors.  
        Location: shortlist regions via priors, then confirm in the most indicative sub-areas.  
        Attribute/State: prioritize proximity checks of the target category using functional/visual cues; verify state locally.  
        Counting/Relationship: gain coverage to enumerate instances first, then verify local relations.  
        Text-reading: seek text-bearing surfaces/signage/panels with high-contrast lettering near decision points (entrances, hubs, boards).  
        Clarify which regions are helpful (connectors such as hallways/intersections, doorways, hubs) and which are harmful (blind dead-ends, purely cluttered corners without new cues).

        \item \textbf{Step 4 (Anti-patterns) --- 2--3 sentences:}  
        Describe common failure modes to avoid.  
        Make it concrete and environment-aware: specify where/when \emph{not} to go. For example: following the perimeter of closed garage doors yields little new evidence when searching for containers; diving into deep storage alcoves is unhelpful for text-reading tasks; lingering in decor-heavy corners seldom helps container/appliance queries; circling vehicle bays rarely reveals recycling signage.  
        Also state when to stop: avoid repeating passes along blank walls or returning to dead-end utility closets after container zones were already scanned; do not switch directions without fresh evidence; treat wrong or full-bin findings as negative evidence to pivot early.
      \end{itemize}

      \item \textbf{ABSTRACTION:}
      \begin{itemize}
        \item \textbf{Abstraction} --- produce a 20--24 sentence cohesive paragraph integrating \textbf{Step 0--4} into actionable, transferable guidance for similar tasks.  
        Do not introduce scope beyond \textbf{Step 0--4}; do not mention BVF/CVF/views/images; do not use step IDs inside the paragraph.  
        The paragraph should condense task understanding, trajectory structure, environment--object associations, question-type strategies, directional priors, and anti-patterns into a single generalized abstraction that can guide future embodied exploration under similar conditions.
      \end{itemize}
    \end{itemize}
\end{itemize}

\textbf{Guidelines:}
\begin{itemize}
  \item Use only region/landmark/path terms and task-relevant cues; do \emph{not} mention cameras, BVF/CVF, views, snapshots, or images.
  \item Keep every block label \emph{exactly} as specified and in order. No extra sections or headers.
  \item Ground all reasoning in the provided task, trajectory caption, and outcome; do not invent scene-specific objects beyond generic environment--object priors.
\end{itemize}

\textbf{Expected Output:}
\begin{itemize}
  \item Write the sections starting with the literal label \texttt{REFLECTION:}, followed by the five blocks \textbf{Step 0--4} in order, and then the section \texttt{ABSTRACTION:} with a single \textbf{Abstraction} paragraph.  
  \item Ensure all reasoning is coherent, causally connected, and directly useful for guiding future trajectories on similar questions.
\end{itemize}

\captionof{figure}{Full Prompt of Retrospective Experience Abstraction.}
\label{sup:abstraction_prompt}
\end{SelfReflectionBox*}
\captionsetup{type=figure}

\begin{SelfReflectionBox*}[title={Hierarchical Frontier Selection}]
\scriptsize     
You are an embodied agent navigating an indoor environment to answer a given question. At each step, you must choose \textbf{exactly one frontier} to explore next.

\begin{itemize}

  \item \textbf{Frontier Definition:}
  \begin{itemize}
    \item \textbf{Broad-View Frontiers (BVF):} Divide your full 360° surroundings into coarse directional sectors. You must pick one BVF to look closer.
    \item \textbf{Close-up-View Frontiers (CVF):} Sub-divisions within the chosen BVF, offering narrower perspectives. You must pick one CVF to move toward.
  \end{itemize}

  \item \textbf{Supporting Contexts:}
  \begin{itemize}

    \item \textbf{Egocentric View (if present):}  
    The agent's immediate forward-looking camera view; use it as local reference only.

    \item \textbf{Episodic Context (if present):}  
    Factual textual summary of the agent's previous steps within this episode (visited path, seen/unseen areas).  
    Use this to avoid redundancy and prefer novel, informative directions.

    \item \textbf{Retrospective Experience / Trajectory Abstraction (if present):}
    \begin{itemize}

      \item \textbf{Retrospective Experience:}  
      Textual memory from a similar question and environment, how a previous decision was made, which frontier was chosen, the resulting outcome, critique, and abstraction.

      \item \textbf{Trajectory Abstraction:}  
        It summarizes region-level transitions, evidence-driven pivots, high-priority areas, low-yield detours, functional cues, and timing logic.  
        Use this abstraction to inform your current decision-making: extract the underlying directional tendencies, preferred region-ordering, and failure-avoidance patterns, and apply them when selecting the next frontier.  
        The abstraction should be interpreted as actionable, environment-grounded strategic knowledge that biases your choices toward informative regions and away from historically unproductive ones, using only regions, landmarks, and functional cues without reference to images, snapshots, or camera operations.

    \end{itemize}

  \end{itemize}

  \item \textbf{Task Rules:}
  \begin{itemize}
    \item You will only be given either BVFs or CVFs at a time.
    \item Your reasoning must be concrete and visual: name specific objects, layouts, textures, lighting, text-bearing surfaces, or spatial cues relevant to the question.
    \item You must select exactly one frontier; do not claim none are suitable.
    \item Output the rationale first, then the final answer on a separate line: \texttt{BVF i} or \texttt{CVF i}.
  \end{itemize}

\end{itemize}

\textbf{Inputs:}
\begin{itemize}
  \item \textbf{Question:} \{question text\}
  \item \textbf{Frontier Candidates:}
  \begin{enumerate}
    \item \textbf{BVF 0:} \{frontier image 0\}
    \item \textbf{BVF 1:} \{frontier image 1\}
    \item \dots
  \end{enumerate}
  \item \textbf{Egocentric Forward View (optional):} \{egocentric image\}
  \item \textbf{Episodic Context (optional):} \{episodic summary\}
  \item \textbf{Retrospective Experience or Trajectory Abstraction (optional):} \{context text\}
\end{itemize}

\textbf{Reasoning Procedure:}
\begin{enumerate}

  \item \textbf{Step 0:}  
  Restate the task in your own words and confirm that you must pick exactly one frontier of the current type.

  \item \textbf{Step 1:}  
  From the \textbf{Episodic Context} (if any), summarize which regions are already explored and which remain unseen.

  \item \textbf{Step 2:}
  \begin{itemize}
    \item If \textbf{Trajectory Abstraction} is present: extract 1–2 concise directive rules that give specific, problem-focused guidance for the current question. 
    \item If \textbf{Retrospective Experience} is present: summarize how previous cases chose, what results followed, and distill 1–2 rules for your current decision.
  \end{itemize}

  \item \textbf{Step 3:}  
  Compare all current frontier candidates one by one using visual cues, novelty, and relevance to the distilled rules; reason toward the best option.

  \item \textbf{FINAL:}  
  Print only the decision line — \texttt{BVF i} or \texttt{CVF i}.

\end{enumerate}

\textbf{Goal:}  
Produce concise, visual, step-wise reasoning that integrates episodic and past-experience cues to choose the most informative frontier.

\captionof{figure}{Full prompt of hierarchical selection exploration.}
\label{sup:frontier_prompt}

\end{SelfReflectionBox*}

\end{document}